\pdfoutput=1

\documentclass[11pt]{article}

\usepackage{anyfontsize}
\usepackage{tipa}
\usepackage{acl}
\usepackage{times}
\usepackage{svg}
\usepackage{latexsym}
\usepackage[T1]{fontenc}
\usepackage[utf8]{inputenc}
\usepackage{array}
\usepackage{hyperref}       
\usepackage{url}            
\usepackage{booktabs}       
\usepackage{amsfonts}       
\usepackage{nicefrac}       
\usepackage{booktabs} 
\usepackage{kotex}
\usepackage{natbib}
\usepackage{arydshln}
\usepackage{amsmath}
\usepackage{lscape}
\usepackage{multirow}
\usepackage{microtype}
\usepackage{graphicx}
\usepackage{enumitem}
\usepackage{makecell}
\usepackage{tabularx}
\usepackage{tcolorbox}
\usepackage{pgfplots}
\usepackage{pgfplotstable}
\usepackage{subcaption}
\usepackage{tipa}
\usepackage{booktabs}
\usepackage[T1]{fontenc} 
\usepackage{pgfplots}
\pgfplotsset{compat=1.18} 
\usepackage{amsmath} 

\usepgfplotslibrary{groupplots}
\usetikzlibrary{patterns}
\pgfplotsset{compat=1.17}

\title{P-CoT: A Pedagogically-motivated Participatory Chain-of-Thought Prompting for Phonological Reasoning in LLMs}

\author{
  Dongjun Jang \and Youngchae Ahn \and Hyopil Shin \\
  Department of Linguistics \\
  Seoul National University \\
  \texttt{\{qwer4107, estelle1026, hpshin\}@snu.ac.kr}
}
\usepackage[T1]{fontenc}  
\usepackage[utf8]{inputenc}  
\usepackage{tipa}  

\begin{document}
\maketitle
\begin{abstract}
This study explores the potential of phonological reasoning within text-based large language models (LLMs). Utilizing the PhonologyBench benchmark, we assess tasks like rhyme word generation, g2p conversion, and syllable counting. Our evaluations across 12 LLMs reveal that while few-shot learning offers inconsistent gains, the introduction of a novel Pedagogically-motivated Participatory Chain-of-Thought (P-CoT) prompt, which is anchored in educational theories like scaffolding and discovery learning, consistently enhances performance. This method leverages structured guidance to activate latent phonological abilities, achieving up to 52\% improvement and even surpassing human baselines in certain tasks. Future work could aim to optimize P-CoT prompts for specific models or explore their application across different linguistic domains.
\end{abstract}

\section{Introduction}
The performance of transformer decoder-based large language models (LLMs), such as ChatGPT \cite{openai2023gpt}, has advanced rapidly in recent years \cite{brown2020language, taylor2022galactica, chowdhery2022palm, shanahan2022talking, touvron2023llama, bubeck2023sparksartificialgeneralintelligence, hu2024gentranslatelargelanguagemodels,  guo2025deepseek}. These models excel at a wide range of tasks through in-context learning \cite{dong2022survey} and have been extensively evaluated for their capabilities in syntax, semantics, morphology, and discourse \cite{liang2022holistic, atox2024evaluating, titus2024does}. In contrast, phonology has received comparatively little attention in this area, primarily because it has traditionally been studied using multi-modal \cite{zhang2023speechgpt, yu2024self} or speech-based models \cite{baevski2020wav2vec, rubenstein2023audiopalm, radford2023robust}. This is despite its significant role in generating natural prosody \cite{zhang2023prosody} and accurately modeling dialectal variations \cite{dutoit1997high}.

Recent work such as \citet{suvarna2024phonologybench} shows that while text-based LLMs encode latent phonological patterns from language data, they often struggle with explicit tasks, resulting in a notable gap compared to human performance. Therefore, enhancing phonological reasoning in text-based LLMs emerges as a timely and important research direction.

\begin{table*}[h]
    \centering
    \begin{tabularx}{\textwidth}{X >{\arraybackslash}X >{\centering\arraybackslash}X}
        \toprule
        \textbf{Task} & \textbf{Example Input} & \textbf{Example Output}\\
        \midrule
        \textbf{Rhyme Word Generation} & Give 5 words that rhyme with `\textit{education}'. & \textit{circulation, occupation, reputation, population, reservation}\\
        \hdashline
        \textbf{G2P Conversion} & Convert the given grapheme \textit{`basement'} into phoneme according to American English in IPA. & \textipa{/beIsm@nt/}\\
        \hdashline
        \textbf{Syllable Counting} & Count the number of syllables in the sentence: "\textit{To top it all off, I miss my stunner.}" & \textit{10}\\
        \bottomrule
    \end{tabularx}
    \caption{Illustrative Example Tasks from PhonologyBench \cite{suvarna2024phonologybench}: Rhyme Word Generation, G2P Conversion, and Syllable Counting}
    \label{tab:phonology_tasks}
\end{table*}

Our study investigates the phonological competence of modern text-based language models using PhonologyBench\footnote{\url{https://github.com/asuvarna31/llm_phonology}} \citep{suvarna2024phonologybench}, a benchmark that evaluates three key tasks: rhyme word generation, grapheme-to-phoneme (g2p) conversion, and syllable counting. We first evaluated 12 state-of-the-art models and examined the impact of few-shot learning \cite{brown2020language} on their performance. Our findings reveal that while few-shot prompting sometimes results in improvements over baseline, these enhancements are inconsistent and model-dependent, with performance occasionally falling below baseline levels. This underscores the need for more robust techniques to enhance phonological reasoning in text-based LLMs.

To address these challenges, we propose a novel approach that integrates role-playing pedagogy within a Chain-of-Thought (CoT) prompting framework \cite{wei2022chain}. We hypothesize that although LLMs inherently encode meta-knowledge in phonology, they lack effective strategies to leverage this information. By integrating two distinct pedagogical techniques—scaffolding \citep{wood1976role} and discovery learning \citep{bruner1961act}—and building on prior studies that have effectively combined these methods \citep{jatisunda2020discovery, alfieri2011does, schofield2021learning}, our Pedagogically-motivated Participatory Chain-of-Thought (P-CoT) prompt consistently improves performance across all 12 models on the phonological tasks. These results persist even though we use the same examples from standard few-shot learning, indicating that our approach, rather than the data itself drives the enhanced performance. Notably, our method achieves performance gains of up to 52\% and even surpasses the human baseline in the specific task. Furthermore, we conducted detailed, task-specific analyses and found that the observed improvements represent true learning advances. These findings underscore the effectiveness of our P-CoT prompts, demonstrating their superiority over traditional few-shot methods in unlocking the latent phonological capabilities of LLMs.

\section{Related Work}

\begin{figure*}[t]
  \centering
  \includegraphics[width=\textwidth]{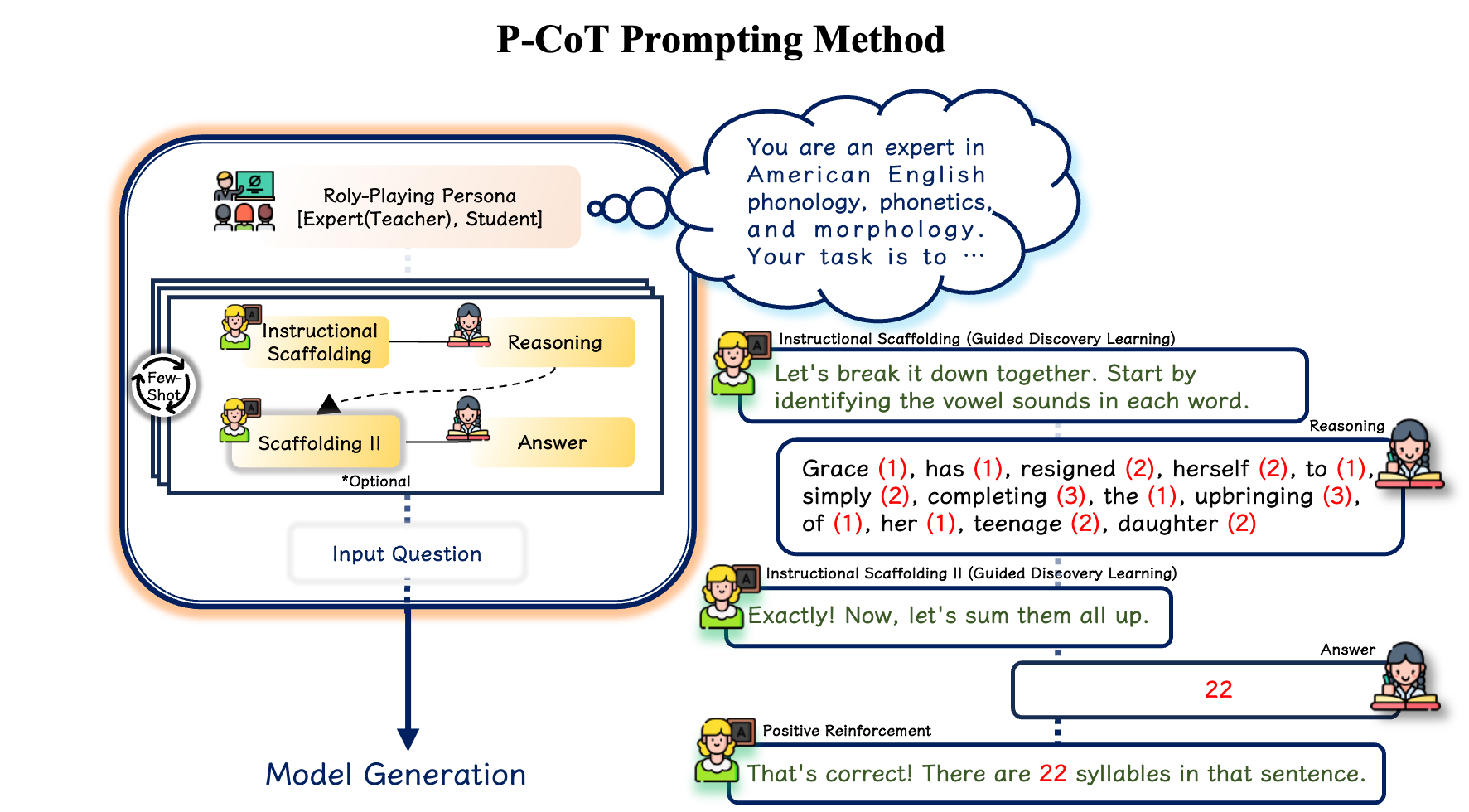}
  \caption{P-CoT Prompting Method}
  \label{fig:speech_production}
\end{figure*}
 
Recent work has begun to explore the phonological skills of text-based LLMs. \citet{suvarna2024phonologybench} introduced PhonologyBench—a benchmark for phonological tasks such as g2p conversion, syllable counting, and rhyme word generation (Table \ref{tab:phonology_tasks}). \citet{qharabagh2024llmpoweredgraphemetophonemeconversionbenchmark} enhanced g2p conversion performance with a two-stage prompting strategy. Additionally, \citet{xue2021deeprapper} proposed DeepRapper, a Transformer-based model that generates rap lyrics, while \citet{loakman2023twistlistresourcesbaselinestongue} introduced the TwistList dataset and benchmarks for tongue twister generation. Together, these studies reveal that research in phonological capabilities is largely task-specific, rather than focusing on a comprehensive evaluation of overall phonological performance.

Alongside these advances, the study of personas in LLMs has progressed significantly, with Role-Playing Language Agents (RPLAs) enhancing human-like performance. \citet{chen2024persona} offer a comprehensive categorization of personas, while \citet{tseng2024two} present a taxonomy of LLM role-playing and personalization. \citet{wang2023unleashing} propose Solo Performance Prompting (SPP), which leverages multiple personas to boost model performance. In parallel, researchers have also turned to educational methods to enhance prompt design. For instance, \citet{chang2023promptinglargelanguagemodels} adopt the Socratic method to guide LLM reasoning through structured questioning, while \citet{jiang2024llms} introduce PedCoT, which uses Bloom’s taxonomy to guide self-correction through a two-stage prompting process involving multiple model calls.

Building on these insights, our approach integrates both persona-based role-play and educational principles to design effective prompts. Specifically, we draw from discovery learning \citep{bruner1961act, anthony1973learning}, where learners explore core principles through inquiry-based methods \citep{van2005co}. Although discovery learning fosters self-directed exploration, studies \citep{moreno2004decreasing, tuovinen1999comparison, hardiman1986learning, brown1994guided, kirschner2006unguided} indicate that structured guidance is crucial to prevent confusion and misconceptions. Given that current LLMs exhibit only rudimentary phonological competence relative to human baselines \citep{suvarna2024phonologybench}, we integrate scaffolding and worked examples into our prompts to provide effective, guided instruction.

Accordingly, we combine discovery learning with the scaffolding technique introduced by \citet{wood1976role} to design our prompts. They define scaffolding as a process that enables a novice to tackle tasks beyond their unaided efforts. Later, \citet{bruner1985vygotsky} and \citet{Cazden1979PeekabooAA} relate this notion to Vygotsky’s Zone of Proximal Development (ZPD)—the space between what learners can do alone and what they can accomplish with guidance. In our approach, scaffolding is used to help the model traverse this zone, offering temporary support that enables it to engage with phonological tasks beyond its current level of competence.

This integration of discovery learning and scaffolding has been shown to be effective in educational contexts. For instance, \citet{ewing2000scaffolding} and \citet{Collins1991COGNITIVEAM} provide real-world examples of scaffolding in educational settings. \citet{Collins1991COGNITIVEAM} particularly outlines a six-phase instructional model centered on modeling, coaching, and scaffolding. Similarly, \citet{jatisunda2020discovery} found that discovery learning with scaffolding improves creative thinking and self-efficacy, outperforming both unguided discovery and conventional methods. Works such as \citet{alfieri2011does} and \citet{schofield2021learning} further support that guided discovery significantly enhances learning outcomes.

\section{Assessment of Phonological Understanding in LLMs}

\begin{table}[!h]
\centering
\begin{tabularx}{0.45\textwidth}{|X|r|}
\hline
\textbf{Feature} & \textbf{Count} \\
\hline
G2P Conversion (High) & 1042 \\
\hline
G2P Conversion (Low) & 2084 \\
\hline
Rhyme Word Generation (Common) & 199 \\
\hline
Rhyme Word Generation (Rare) & 110 \\
\hline
Syllable Counting & 993 \\
\hline
\end{tabularx}
\caption{Distribution of PhonologyBench Data} \
\label{tab:phonologybench}
\end{table}

In this section, we aim to explore the extent to which LLMs inherently possess phonological knowledge by evaluating their performance in a zero-shot context. Additionally, we plan to examine the effectiveness of few-shot learning in further enhancing the models' phonological reasoning. In our exploration of the phonological capabilities of text-based LLMs, we focus on evaluating three key tasks: g2p conversion, rhyme word generation, and syllable counting. We utilize the PhonologyBench framework \citet{suvarna2024phonologybench} as a comprehensive benchmark for assessing these tasks.

\begin{table*}[t]
\centering
\begin{tabularx}{\textwidth}{|l|c|c|c|c|c|c|}
\hline
Model(Open/Closed) & Baseline & 3-Shot & 5-Shot & \textbf{P-CoT1} & \textbf{P-CoT3} & \textbf{P-CoT5} \\
\hline
Llama-3.3-70B-Instruct & 66.3/34.4 & 65.3/38.4 & 64.7/36.9 & \textbf{\textcolor{black}{77.2}}/47.1 & 76.5/46.4 & 76.4/\textbf{\textcolor{black}{47.5}} \\
Llama-3.1-8B-Instruct & 41.7/17.5 & 41.3/21.6 & 39.7/27.1 & 54.7/30.2 & 55.5/33.3 & \textbf{\textcolor{black}{62.9}}/\textbf{\textcolor{black}{37.1}} \\
Mistral-7B-Instruct-v0.2 & 26.8/12.0 & 28.5/9.3 & 27.9/9.3 & 31.9/15.2 & 71.6/30.2 & \textbf{\textcolor{black}{78.8}}/\textbf{\textcolor{black}{36.2}} \\
Ministral-8B-Instruct-2410 & 16.2/8.7 & 26.6/6.9 & 28.6/8.9 & 34.4/20.8 & 50.2/22.7 & \textbf{\textcolor{black}{63.9}}/\textbf{\textcolor{black}{31.9}} \\
Qwen2.5-72B-Instruct & 51.8/26.7 & 54.8/25.8 & 53.6/22.4 & 66.6/35.8 & 68.1/\textbf{\textcolor{black}{36.2}} & \textbf{\textcolor{black}{69.9}}/35.4 \\
Qwen2.5-7B-Instruct & 23.7/9.5 & 21.2/7.3 & 22.1/6.7 & 43.5/15.6 & 45.8/29.2 & \textbf{\textcolor{black}{48.5}}/\textbf{\textcolor{black}{30.2}} \\
gemma-2-9b-it & 55.6/40.5 & 57.1/38.7 & 57.6/39.3 & \textbf{\textcolor{black}{73.0}}/52.5 & 71.8/47.7 & 71.2/\textbf{\textcolor{black}{62.3}} \\
gemma-2-27b-it & 62.4/39.6 & 64.5/34.9 & 63.4/34.7 & \textbf{\textcolor{black}{78.4}}/\textbf{\textcolor{black}{52.7}} & 77.3/51.3 & 77.7/51.2 \\
\hdashline
gpt-3.5-turbo & 70.7/47.3 & 72.9/45.1 & 73.1/45.5 & 79.2/\textbf{\textcolor{black}{61.9}} & 80.5/59.2 & \textbf{\textcolor{black}{81.3}}/51.2 \\
gpt-4o & 73.0/50.7 & 73.8/44.9 & 74.7/49.5 & - & \textbf{\textcolor{black}{86.0}}/\textbf{\textcolor{black}{60}} & 84.3/57.5 \\
Claude 3.5 Sonnet & 76.4/38.2 & 76.9/36.7 & 77.3/37.1 & 82.5/50.8 & 83.1/\textbf{\textcolor{black}{53.3}} & \textbf{\textcolor{black}{83.2}}/51.2 \\
Claude 3.5 Haiku & 68.5/41.5 & 70.2/41.1 & 70.5/41.5 & 80.2/34.8 & 81.7/58.7 & \textbf{\textcolor{black}{82.4}}/\textbf{\textcolor{black}{58.8}} \\
\hline
Human Baseline & \multicolumn{6}{c|}{86.4/60.4} \\
\hline
\end{tabularx}
\caption{Performance Comparison on Rhyme Word Generation Tasks (Common/Rare Word Sets)}
\label{tab:performance_comparison_rhyme}
\end{table*}

\subsection{Phonological Benchmark Definition}

We adopted the PhonologyBench framework \citep{suvarna2024phonologybench}, which forms a critical component of our study, to systematically evaluate the phonological knowledge of LLMs. Specifically, it includes datasets with 3,126 words for g2p conversion, categorized into 2,084 high-frequency and 1,042 low-frequency words. Additionally, it offers a dataset of 309 words, split into 199 common words and 110 rare ones for rhyme word generation tasks. For each word, a gold reference set was constructed by matching, on average, 1,200 rhyming candidates, ensuring a robust and comprehensive evaluation criterion. For syllable counting, the benchmark includes 993 sentences (Table \ref{tab:phonologybench}). 

\subsection{Models}
We aim to evaluate the overall performance of LLMs on typical phonological tasks by assessing the performance of up to 12 different LLM models. Among them, 8 are open-source models and 4 are closed models. 

\paragraph{Open Models} We select a diverse set of open-source models, including both foundational and advanced versions of the latest LLMs, with parameter counts ranging from 7 billion to 72 billion. This lineup features Meta's LLaMA series models like Llama-3.3-70B and Llama-3.1-8B, as well as Mistral's Mistral-7B and Ministral-8B \citep{dubey2024llama, jiang2023mistral, Ministral8BInstruct2410}. Additionally, we include various iterations from the Qwen and gemma series \citep{yang2024qwen2,team2024gemma}.

\begin{table*}[t]
\centering
\begin{tabularx}{\textwidth}{|l|c|c|c|c|c|c|}
\hline
Model (Open/Closed) & Baseline & 3-Shot & 5-Shot & \textbf{P-CoT1} & \textbf{P-CoT3} & \textbf{P-CoT5} \\
\hline
Llama-3.3-70B-Instruct & 27.0/40.5 & 29.9/42.9 & 29.6/43.8 & 32.2/45.6 & 32.8/48.8 & \textbf{\textcolor{black}{33.2}}/\textbf{\textcolor{black}{49.7}} \\
Llama-3.1-8B-Instruct & 17.2/24.1 & 21.3/31.5 & 21.4/31.9 & 21.8/\textbf{\textcolor{black}{31.9}} & \textbf{\textcolor{black}{22.4}}/29.1 & 22.0/29.8 \\
Mistral-7B-Instruct-v0.2 & 9.5/15.0 & 13.5/19.2 & 13.1/19.3 & 14.3/22.2 & \textbf{\textcolor{black}{16}}/21.9 & 15.6/\textbf{\textcolor{black}{22.4}} \\
Ministral-8B-Instruct-2410 & 19.9/31.4 & 20.3/30.8 & 20.6/32.1 & 22.3/32.5 & \textbf{\textcolor{black}{23.7}}/\textbf{\textcolor{black}{35.3}} & 23.1/34.6 \\
Qwen2.5-72B-Instruct & 25.3/38.0 & 30.0/42.1 & 28.7/43.2 & 28.2/40.2 & \textbf{\textcolor{black}{30.1}}/\textbf{\textcolor{black}{43.4}} & 28.8/41.4 \\
Qwen2.5-7B-Instruct & 12.5/18.6 & 13.3/19.9 & \textbf{\textcolor{black}{13.6}}/\textbf{\textcolor{black}{22.3}} & 10.0/17.4 & 10.8/17.9 & 10.2/16.6 \\
gemma-2-9b-it & 24.1/36.2 & 25.9/40.0 & 26.4/39.8 & 27.7/40.5 & \textbf{\textcolor{black}{28.6}}/\textbf{\textcolor{black}{44.2}} & 27.4/42.3 \\
gemma-2-27b-it & 29.6/43.8 & 30.4/46.9 & 31.1/46.6 & 31.1/46.7 & \textbf{\textcolor{black}{32.1}}/\textbf{\textcolor{black}{48.7}} & 31.4/47.9 \\
\hdashline
gpt-3.5-turbo & 28.9/43.2 & 42.6/60.3 & 38.7/56.8 & 45.4/61.9 & \textbf{\textcolor{black}{50.1}}/\textbf{\textcolor{black}{68.3}} & 49.3/67.5 \\
gpt-4o & 32.0/49.8 & 46.9/69.5 & 44.1/57.1 & 49.7/69.2 & 51.7/\textbf{\textcolor{black}{69.8}} & \textbf{\textcolor{black}{52.1}}/69.6 \\
Claude 3.5 Sonnet & 35.5/51.6 & 57.2/79.3 & 57.0/79.8 & 59.9/80.4 & 61.0/81.3 & \textbf{\textcolor{black}{61.6}}/\textbf{\textcolor{black}{82.2}} \\
Claude 3.5 Haiku & 33.5/49.3 & 55.9/\textbf{\textcolor{black}{76.4}} & 54.4/74.5 & 54.0/74.5 & \textbf{\textcolor{black}{56.9}}/75.1 & 56.5/75.4 \\
\hline
\end{tabularx}
\caption{Performance Comparison on G2P Conversion Tasks (low/high Word Sets)}
\label{tab:performance_comparison_g2p}
\end{table*}

\paragraph{Closed Models} In addition to the open-source models, we also examine a selection of top-tier, proprietary models, which include OpenAI's GPT series (featuring GPT-3.5-turbo and GPT-4o \citep{openaiGPT3.5turbo, gpt4all}) and Anthropic's Claude series (Claude 3.5 Sonnet and Claude 3.5 Haiku \citep{anthropic2024claude, anthropicClaude3.5Haiku}).

\subsection{Metrics for Model Evaluation}
To assess the performance of the models on g2p conversion and syllable counting tasks, we use the Exact Match criterion as our measure of accuracy. For the rhyme word generation task, we evaluate performance using a Success Rate metric. This metric calculates the proportion of rhyming words generated by the model that match those in a predefined set of expected outcomes. Specifically, we determine the success rate for each word by finding the ratio of correctly generated candidates present in the ground-truth set. The overall Success Rate (SR) is then derived as the average of these individual success rates across all words. By applying these metrics, we remain consistent with the evaluation methods used in PhonologyBench \citep{suvarna2024phonologybench}.

\subsection{Configuration}
To ensure consistent and reproducible results across our experiments, we fix the model generation probability by setting a deterministic seed. This allows us to maintain consistent outputs across different trials and accurately evaluate the true performance improvements attributed to our methods. We conduct inference using several A100 GPUs, each equipped with 80GB of memory, without applying techniques like quantization \citep{frantar2022gptq, liu2023llm} or FlashAttention \citep{dao2023flashattention}. Also, we adhere closely to the zero-shot prompts provided by PhonologyBench to accurately assess the models.

\section{Results and Analysis}
\subsection{Analysis of Few-Shot Learning Effectiveness}
We analyzed the effectiveness of few-shot learning in phonological reasoning tasks. We compared this approach to traditional baseline models across various LLMs and phonological tasks to assess performance differences and identify any notable improvements.
\paragraph{\textbf{Rhyme Word Generation:}} 
In the rhyme word generation task, few-shot learning demonstrated mixed results, as shown in Table \ref{tab:performance_comparison_rhyme}. While some models, like GPT-3.5-turbo, showed improvements over the baseline with a 2-3 percentage point increase in accuracy for common rhymes, others, such as the Llama, Qwen, and gemma series, experienced only marginal gains or even declines. This inconsistency underscores the varying ability of models to leverage contextual examples efficiently, indicating that the effectiveness of few-shot learning is heavily dependent on the model's intrinsic capacity to process and learn from examples. 

\paragraph{\textbf{G2P Conversion:}} In the g2p conversion task, as shown in Table \ref{tab:performance_comparison_g2p}, few-shot learning led to variable results. Models like Claude 3.5 Sonnet experienced a significant boost, increasing from a baseline of 35.5\% to 57.2\% accuracy on the low-frequency set. Conversely, other models, such as gemma-2-27b-it, showed minimal progress, remaining near their baseline scores. 

\paragraph{\textbf{Syllable Counting:}} 

As presented in Table \ref{tab:performance_comparison_syllable}, few-shot learning provided modest gains in the syllable counting task. Notably, models like Claude 3.5 Haiku achieved a significant improvement, rising from a baseline accuracy of 21.1\% to 57.4\%. However, others, such as Mistral-7B and Ministral-8B, showed little advancement. 
\begin{table*}[t]
\centering
\begin{tabular}{|l|c|c|c|c|c|c|}
\hline
Model (Open/Closed) & Baseline & 3-Shot & 5-Shot & \textbf{{P-CoT1}} & \textbf{{P-CoT3}} & \textbf{{P-CoT5}} \\
\hline
Llama-3.3-70B-Instruct & 18.6 & 13.5 & 11.6 & \textbf{\textcolor{black}{41.6}} & 28.7 & 28.1 \\
Llama-3.1-8B-Instruct & 13.0 & 14.2 & 13.3 & \textbf{\textcolor{black}{41.3}} & 37.4 & 39.6 \\
Mistral-7B-Instruct-v0.2 & 2.9 & 3.7 & 4.4 & 13.0 & \textbf{\textcolor{black}{26.5}} & 14.9 \\
Ministral-8B-Instruct-2410 & 15.6 & 3.9 & 5.8 & 10.8 & \textbf{\textcolor{black}{26.5}} & 14.9 \\
Qwen2.5-72B-Instruct & 12.9 & 14.6 & 15.0 & 14.8 & \textbf{\textcolor{black}{15.6}} & 10.7 \\
Qwen2.5-7B-Instruct & 1.0 & 2.1 & 4.8 & 2.5 & \textbf{\textcolor{black}{10.3}} & 4.3 \\
gemma-2-9b-it & 7.3 & 9.3 & 8.7 & \textbf{\textcolor{black}{19.4}} & 8.7 & 12.6 \\
gemma-2-27b-it & 13.6 & 23.4 & \textbf{\textcolor{black}{23.5}} & 17.1 & 20.3 & 20.6 \\
\hdashline
gpt-3.5-turbo & 16.0 & 19.1 & 19.1 & \textbf{\textcolor{black}{48.8}} & 46.5 & 42.8 \\
gpt-4o & 20.8 & 20.9 & 19.5 & 20.4 & \textbf{\textcolor{black}{22.2}} & 19.8 \\
Claude 3.5 Sonnet & 20.2 & 24.8 & \textbf{\textcolor{black}{32.9}} & 23.4 & 28.8 & 32.6 \\
Claude 3.5 Haiku & 21.1 & 20.7 & 26.9 & 48.2 & \textbf{\textcolor{black}{57.4}} & 54.5 \\
\hline
\end{tabular}
\caption{Performance Comparison on Syllable Counting Tasks}
\label{tab:performance_comparison_syllable}
\end{table*}

\subsection{P-CoT: A Pedagogically-motivated Participatory Chain of Thought Prompting}\footnote{\url{https://github.com/Junmaij/P-CoT}}
Building on our initial observations that revealed limited success with traditional few-shot learning in enhancing phonological reasoning, we propose a novel approach: the Pedagogically-motivated Participatory Chain-of-Thought (P-CoT). This framework is specifically designed to expand the latent phonological capabilities inherent in text-based LLMs (Figure \ref{fig:speech_production}). Our method integrates discovery learning \citep{bruner1961act, anthony1973learning}, and scaffolding \citep{wood1976role} to provide structured guidance and contextual support. Additionally, inspired by Vygotsky’s ZPD \citep{vygotsky1978mind} and the instructional strategies of \citet{Collins1991COGNITIVEAM} and \citet{jatisunda2020discovery}, we offer temporary reciprocal guidance through targeted hints, clear definitions, and specific tasks. Moreover, by emphasizing the active engagement of the learner, our prompt design ensures that attentive and motivated models effectively navigate the discovery process and avoid misconceptions, ultimately improving performance on phonological tasks. 

Our approach actively engages LLMs with structured prompts that simulate guided exploration, leveraging their capacity to infer and generalize from examples, as previous educational research has recommended \citep{tuovinen1999comparison, moreno2004decreasing}. This also facilitates their internalization of complex phonological rules and is expected to surpass the capabilities of traditional few-shot learning \citep{brown2020language}, enabling a more intricate application of phonological knowledge.

\subsubsection{Prompt Design}
In our exploration of the P-CoT methodology, we utilized the same sample sets used in our earlier few-shot learning experiments to ensure consistency in evaluating the models' performance enhancements. This approach allowed us to directly compare the efficacy of our P-CoT approach against the baselines established in previous tests. 

Specifically, we defined three distinct configurations based on the number of interactions or dialogues included in the P-CoT prompt:

\begin{tcolorbox}[colback=blue!8!white,colframe=blue!60!black,title=P-CoT Prompt,fonttitle=\bfseries\small]
\begin{itemize}
    \item \textbf{P-CoT1:} Presents a single teacher-student interaction, where the model is guided to derive a phonological rule from a single example.
    \item \textbf{P-CoT3:} Extends the same instructional format to three examples, while maintaining the dialogic structure.
    \item \textbf{P-CoT5:} Further increases the number of interactions to five, reinforcing the target concept through guided discovery learning.
\end{itemize}
\end{tcolorbox}

The general structure of our P-CoT prompts is designed to simulate an instructional setting grounded in educational theory. Specifically, the prompts implement a form of guided discovery learning, which combines the principles of discovery learning with scaffolding techniques that support the learner's reasoning process. Each prompt begins by assigning teacher and student personas to the model and the user, respectively, thereby establishing an interactive and participatory learning environment. This role-based setup reflects a classroom-like context. 

To initiate this interaction, the student persona is either explicitly signaled by phrases like “I'm ready to learn about \textasciitilde,” or implicitly conveyed through contextual cues and role-setting instructions. This design is inspired by \citet{jatisunda2020discovery} and \citet{wood1976role}, who emphasize that guided discovery learning requires learners to actively pursue knowledge. Following this setup, the teacher persona introduces the conceptual basis of the task, providing necessary definitions to prevent misconceptions—serving as the first scaffolding step \citep{jatisunda2020discovery}. Then, consistent with \citet{schofield2021learning}, our prompt design does not aim to convey abstract concepts directly. Instead, the teacher persona guides the student role toward understanding by prompting it to solve concrete problems. To further support this learning process, we apply scaffolding by decomposing the problem into subtasks—a core principle emphasized by \citet{Collins1991COGNITIVEAM} and \citet{ewing2000scaffolding}. For instance, in the syllable counting task, the teacher presents a supporting subtask, prompting the student to identify the vowel sounds in each word before proceeding to count the total syllables in the given sentence. 

As discussed above, we embed a teacher-student dialogue that simulates an instructional exchange into the prompt, aiming to enhance the model’s reasoning ability by situating it within a step-by-step learning process. By applying these structured prompts designed around educational theories of discovery learning and scaffolding, we aim to comprehensively assess the impact of P-CoT on improving phonological reasoning over traditional few-shot learning methods. The detailed design of these prompts is provided in the Appendix \ref{sec:appendix}.

\subsubsection{Result}

The implementation of P-CoT prompts consistently improved performance over the baseline across all examined models and tasks. By simulating a guided discovery learning environment, this approach capitalized on structured exploration and the deliberate scaffolding of knowledge, thereby facilitating deeper cognitive engagement and enhancing phonological competence.

\paragraph{Rhyme Word Generation:} In the domain of rhyme word generation, the P-CoT strategy significantly enhanced model performance, as shown in Table \ref{tab:performance_comparison_rhyme}. The P-CoT approach leveraged structured prompts to effectively improve the model's ability to generate rhymes, achieving notable advancements in both common and rare word categories. For instance, models such as Mistral-7B-Instruct-v0.2 and Ministral-8B-Instruct-2410 demonstrated performance increases of approximately 47.7 to 52 percentage points over the baseline when tasked with common rhymes.  Even within closed models like GPT-4, structured pedagogical approaches led to performance levels approaching that of human baseline. 

\paragraph{G2P Conversion:} In the g2p conversion, the P-CoT approach significantly boosted model performance, as evidenced in Table \ref{tab:performance_comparison_g2p}. By utilizing structured prompts that guide learning, we observed substantial improvements in accuracy across both high-frequency and low-frequency word categories. Notably, models like Claude 3.5 Sonnet experienced a marked increase in successful conversions, with accuracy improvements reaching up to 30 percentage points above baseline levels. 

\paragraph{Syllable Counting:} In the syllable counting task, the P-CoT method showed notable progress, significantly enhancing the performance of multiple models, as detailed in Table \ref{tab:performance_comparison_syllable}. For instance, GPT-3.5-turbo saw its accuracy leap from a baseline of 16.0\% to 48.8\% under the P-CoT1 configuration. Similarly, Claude 3.5 Haiku improved from 21.1\% to 57.4\%, illustrating the potency of structured learning prompts.

\section{Evaluating the Pedagogical Impact on Phonological Comprehension: A Deep Dive into P-CoT Efficacy}
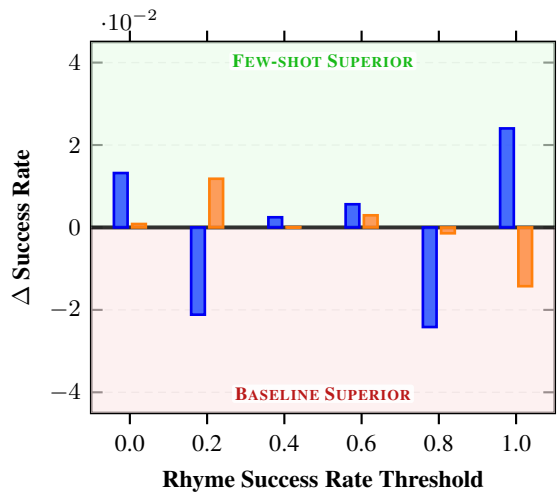
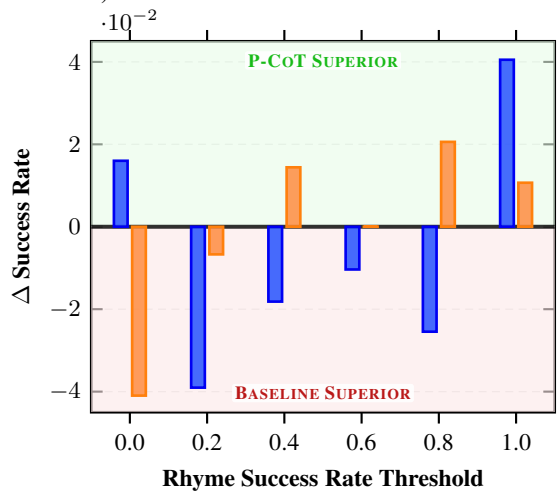
\begin{figure}[!h]
\centering

\begin{subfigure}[b]{0.48\textwidth}
\centering
\begin{tikzpicture}
\begin{axis}[
    width=\textwidth,
    height=6.5cm,
    ybar,
    bar width=0.18cm,
    ymajorgrids=true,
    grid style={densely dashed, gray!25, line width=0.5pt},
    xlabel={Rhyme Success Rate Threshold},
    xlabel style={font=\small\bfseries},
    ylabel={$\Delta$ Success Rate},
    ylabel style={font=\small\bfseries, align=center},
    title style={font=\large\bfseries, yshift=10pt},
    xticklabels={0.0, 0.2, 0.4, 0.6, 0.8, 1.0},
    xtick={0,1,2,3,4,5},
    x tick label style={font=\footnotesize},
    xmin=-0.5,
    xmax=5.5,
    ymin=-0.045,
    ymax=0.045,
    ytick={-0.04,-0.02,0,0.02,0.04},
    yticklabel style={
        font=\footnotesize,
        /pgf/number format/fixed,
        /pgf/number format/precision=2
    },
    axis background/.style={fill=gray!2},
    axis line style={line width=1pt, black},
    tick style={line width=0.8pt, black},
    every axis plot/.append style={thick},
]

\fill[green!8, opacity=0.6] (axis cs:-0.5,0) rectangle (axis cs:5.5,0.045);
\fill[red!8, opacity=0.6] (axis cs:-0.5,-0.045) rectangle (axis cs:5.5,0);

\draw[black, line width=1.5pt, opacity=0.8] (axis cs:-0.5,0) -- (axis cs:5.5,0);

\addplot[
    ybar,
    fill=blue!80!cyan!85,
    draw=blue!95!black,
    line width=1pt,
    bar shift=-0.12cm,
    fill opacity=0.85,
    draw opacity=1,
] coordinates {
    (0, 0.01320038)
    (1, -0.02116431)
    (2, 0.0024692)
    (3, 0.00562781)
    (4, -0.02416135)
    (5, 0.02402827)
};

\addplot[
    ybar,
    fill=orange!80!red!75,
    draw=orange!95!red!95,
    line width=1pt,
    bar shift=0.12cm,
    fill opacity=0.85,
    draw opacity=1,
] coordinates {
    (0, 0.000838250306)
    (1, 0.01181220)
    (2, 0.0000767634)
    (3, 0.00294746)
    (4, -0.00141757)
    (5, -0.01425711)
};

\node[anchor=south, font=\scriptsize\bfseries, text=green!70!black,
      fill=white, rounded corners=1pt, inner sep=1pt, opacity=0.9] 
    at (axis cs:2.5,0.038) {\textsc{Few-shot Superior}};
\node[anchor=north, font=\scriptsize\bfseries, text=red!70!black,
      fill=white, rounded corners=1pt, inner sep=1pt, opacity=0.9] 
    at (axis cs:2.5,-0.038) {\textsc{Baseline Superior}};

\end{axis}
\end{tikzpicture}
\caption{Few-shot over Baseline (Blue: Common Words, Orange: Rare Words)}
\label{fig:rhyme_analysis_few}
\end{subfigure}
\hfill
\begin{subfigure}[b]{0.48\textwidth}
\centering
\begin{tikzpicture}
\begin{axis}[
    width=\textwidth,
    height=6.5cm,
    ybar,
    bar width=0.18cm,
    ymajorgrids=true,
    grid style={densely dashed, gray!25, line width=0.5pt},
    xlabel={Rhyme Success Rate Threshold},
    xlabel style={font=\small\bfseries},
    ylabel={$\Delta$ Success Rate},
    ylabel style={font=\small\bfseries, align=center},
    title style={font=\large\bfseries, yshift=10pt},
    xticklabels={0.0, 0.2, 0.4, 0.6, 0.8, 1.0},
    xtick={0,1,2,3,4,5},
    x tick label style={font=\footnotesize},
    xmin=-0.5,
    xmax=5.5,
    ymin=-0.045,
    ymax=0.045,
    ytick={-0.04,-0.02,0,0.02,0.04},
    yticklabel style={
        font=\footnotesize,
        /pgf/number format/fixed,
        /pgf/number format/precision=2
    },
    axis background/.style={fill=gray!2},
    axis line style={line width=1pt, black},
    tick style={line width=0.8pt, black},
    every axis plot/.append style={thick},
]

\fill[green!8, opacity=0.6] (axis cs:-0.5,0) rectangle (axis cs:5.5,0.045);
\fill[red!8, opacity=0.6] (axis cs:-0.5,-0.045) rectangle (axis cs:5.5,0);

\draw[black, line width=1.5pt, opacity=0.8] (axis cs:-0.5,0) -- (axis cs:5.5,0);

\addplot[
    ybar,
    fill=blue!80!cyan!85,
    draw=blue!95!black,
    line width=1pt,
    bar shift=-0.12cm,
    fill opacity=0.85,
    draw opacity=1,
] coordinates {
    (0, 0.01600783)
    (1, -0.03903264)
    (2, -0.01815913)
    (3, -0.01036432)
    (4, -0.02545858)
    (5, 0.04052315)
};

\addplot[
    ybar,
    fill=orange!80!red!75,
    draw=orange!95!red!95,
    line width=1pt,
    bar shift=0.12cm,
    fill opacity=0.85,
    draw opacity=1,
] coordinates {
    (0, -0.04096768)
    (1, -0.00670211)
    (2, 0.01442308)
    (3, 0.00003267)
    (4, 0.02059643)
    (5, 0.01068513)
};

\node[anchor=south, font=\scriptsize\bfseries, text=green!70!black,
      fill=white, rounded corners=1pt, inner sep=1pt, opacity=0.9] 
    at (axis cs:2.5,0.038) {\textsc{P-CoT Superior}};
\node[anchor=north, font=\scriptsize\bfseries, text=red!70!black,
      fill=white, rounded corners=1pt, inner sep=1pt, opacity=0.9] 
    at (axis cs:2.5,-0.038) {\textsc{Baseline Superior}};

\end{axis}
\end{tikzpicture}
\caption{P-CoT over Baseline (Blue: Common Words, Orange: Rare Words)}
\label{fig:rhyme_analysis_Peda-CoT}
\end{subfigure}

\caption{Success Rate Improvement of Rhyme Word Generation Analysis}
\label{fig:combined_rhyme_analysis}
\end{figure}
In our exploration of enhancing phonological reasoning within text-based LLMs through the P-CoT methodology, our results underscore the significant potential of educationally-inspired frameworks to enhance model performance across all three evaluated phonological tasks: rhyme word generation, g2p conversion, and syllable counting.

\subsection{Rhyme Word Generation}

In Figure \ref{fig:combined_rhyme_analysis}, the performance of two instructional strategies—few-shot and P-CoT—against the baseline is compared for rhyme word generation. Both results illustrate the success rate differences across six different conditions (success rate thresholds from '0.0' to '1.0'), with the y-axis explicitly representing the performance difference between the strategies and the baseline.

For common words, few-shot displays instability, even falling below baseline at higher success rate thresholds like \textit{0.8}. In contrast, P-CoT steadily improves over the baseline, notably at \textit{1.0} thresholds, demonstrating its effectiveness through structured prompts that consistently enhance model performance.

For rare words, few-shot shows limited gains and often underperforms relative to the baseline, particularly at \textit{1.0}, while P-CoT maintains positive improvements at \textit{1.0} success rate. This consistency highlights P-CoT's reliable handling of complex word sets, emphasizing its robustness over few-shot. Thus, P-CoT stands out as the more effective strategy for achieving high reliability and performance in generating rhymes, making it a preferable approach for both common and rare word categories.
\subsection{G2P Conversion}
\begin{figure}[!h]
\centering

\begin{subfigure}[b]{0.48\textwidth}
\centering
\begin{tikzpicture}
\begin{axis}[
    width=\textwidth,
    height=6.5cm,
    ybar,
    bar width=0.15cm,
    ymajorgrids=true,
    grid style={densely dashed, gray!25, line width=0.5pt},
    xlabel={Complexity Score Range},
    xlabel style={font=\small\bfseries},
    ylabel={Accuracy (\%)},
    ylabel style={font=\small\bfseries, align=center},
    title style={font=\large\bfseries, yshift=10pt},
    xticklabels={2.1-3.5, 3.5-4.2, 4.2-5.6, 5.6-6.3, 6.3-14.7},
    xtick={0,1,2,3,4},
    x tick label style={font=\tiny, rotate=0},
    xmin=-0.5,
    xmax=4.5,
    ymin=0,
    ymax=60,
    ytick={0,20,40,60},
    yticklabel style={
        font=\footnotesize,
        /pgf/number format/fixed,
        /pgf/number format/precision=0
    },
    legend style={
        at={(0.5,0.85)},
        anchor=south,
        legend columns=-1,
        draw=black, 
        line width=0.5pt,
        fill=white,
        font=\scriptsize,
        inner sep=3pt,
        rounded corners=1pt,
    },
    axis background/.style={fill=gray!2},
    axis line style={line width=1pt, black},
    tick style={line width=0.8pt, black},
    every axis plot/.append style={thick},
]

\addplot[
    ybar,
    fill=red!70!black!20,
    draw=red!80!black,
    line width=1pt,
    bar shift=-0.2cm,
    fill opacity=0.85,
    draw opacity=1,
] coordinates {
    (0, 42.6)
    (1, 29.5)
    (2, 21.4)
    (3, 16.1)
    (4, 10.6)
};

\addplot[
    ybar,
    fill=blue!70!cyan!30,
    draw=blue!80!black,
    line width=1pt,
    bar shift=0cm,
    fill opacity=0.85,
    draw opacity=1,
] coordinates {
    (0, 48.6)
    (1, 36.8)
    (2, 29.0)
    (3, 21.6)
    (4, 15.0)
};

\addplot[
    ybar,
    fill=green!60!black!20,
    draw=green!80!black,
    line width=1pt,
    bar shift=0.2cm,
    fill opacity=0.85,
    draw opacity=1,
] coordinates {
    (0, 51.4)
    (1, 39.0)
    (2, 31.7)
    (3, 23.5)
    (4, 16.5)
};
\legend{Baseline, Few-shot, P-CoT}
\end{axis}
\end{tikzpicture}
\caption{Low-Frequency Words}
\label{fig:g2p_low_freq}
\end{subfigure}
\hfill
\begin{subfigure}[b]{0.48\textwidth}
\centering
\begin{tikzpicture}
\begin{axis}[
    width=\textwidth,
    height=6.5cm,
    ybar,
    bar width=0.15cm,
    ymajorgrids=true,
    grid style={densely dashed, gray!25, line width=0.5pt},
    xlabel={Complexity Score Range},
    xlabel style={font=\small\bfseries},
    ylabel={Accuracy (\%)},
    ylabel style={font=\small\bfseries, align=center},
    title style={font=\large\bfseries, yshift=10pt},
    xticklabels={1.4-3.5, 3.5-4.2, 4.2-4.9, 4.9-5.6, 5.6-9.8},
    xtick={0,1,2,3,4},
    x tick label style={font=\tiny, rotate=0},
    xmin=-0.5,
    xmax=4.5,
    ymin=0,
    ymax=70,
    ytick={0,20,40,60},
    yticklabel style={
        font=\footnotesize,
        /pgf/number format/fixed,
        /pgf/number format/precision=0
    },
    legend style={
        at={(0.5,0.85)},
        anchor=south,
        legend columns=-1,
        draw=black, 
        line width=0.5pt,
        fill=white,
        font=\scriptsize,
        inner sep=3pt,
        rounded corners=1pt,
    },
    axis background/.style={fill=gray!2},
    axis line style={line width=1pt, black},
    tick style={line width=0.8pt, black},
    every axis plot/.append style={thick},
]

\addplot[
    ybar,
    fill=red!70!black!20,
    draw=red!80!black,
    line width=1pt,
    bar shift=-0.2cm,
    fill opacity=0.85,
    draw opacity=1,
] coordinates {
    (0, 52.5)
    (1, 36.1)
    (2, 33.2)
    (3, 33.7)
    (4, 20.4)
};

\addplot[
    ybar,
    fill=blue!70!cyan!30,
    draw=blue!80!black,
    line width=1pt,
    bar shift=0cm,
    fill opacity=0.85,
    draw opacity=1,
] coordinates {
    (0, 57.6)
    (1, 42.8)
    (2, 39.9)
    (3, 39.6)
    (4, 25.1)
};

\addplot[
    ybar,
    fill=green!60!black!20,
    draw=green!80!black,
    line width=1pt,
    bar shift=0.2cm,
    fill opacity=0.85,
    draw opacity=1,
] coordinates {
    (0, 61.5)
    (1, 49.4)
    (2, 43.8)
    (3, 44.1)
    (4, 29.7)
};

\legend{Baseline, Few-shot, P-CoT}

\end{axis}
\end{tikzpicture}
\caption{High-Frequency Words}
\label{fig:g2p_high_freq}
\end{subfigure}

\caption{Accuracy by Word Complexity of G2P Conversion (Red: Baseline, Blue: Few-shot, Green: P-CoT)}
\label{fig:g2p_conversion}
\end{figure}
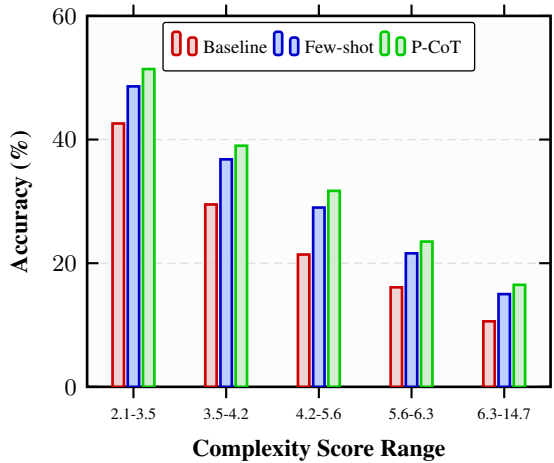
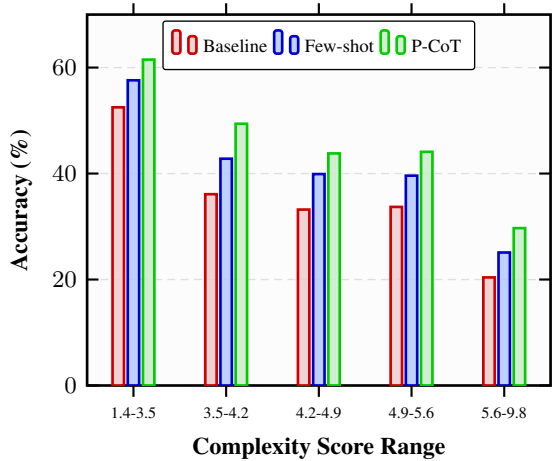

To better understand the result in Table \ref{tab:performance_comparison_g2p}, we analyze model performance against a complexity score \( S \) derived from the formula \( S = 0.4L + 0.3V + 0.3C \), where \( L \) indicates word length, \( V \) counts vowels, and \( C \) assesses consonant quantity. Figures \ref{fig:g2p_low_freq} and \ref{fig:g2p_high_freq} illustrate a clear trend: as word complexity rises, accuracy tends to decline across all evaluated methods, including baseline, few-shot, and P-CoT. However, P-CoT consistently outshines the other methods, even as complexity increases, demonstrating its robust capability to handle intricate g2p conversion.

For words with lower complexity, all approaches achieve relatively high accuracy. It is at higher levels of complexity that the performance gap becomes more pronounced, with P-CoT showing considerable advances. This effectiveness is further confirmed by Table \ref{tab:pvalues}, where we see statistically significant improvements of P-CoT over baseline models in both types of words. Consequently, P-CoT clearly emerges as a superior technique for enhancing phonological reasoning in complex scenarios.

\begin{table}[h]
    \centering
    \begin{tabularx}{0.45\textwidth}{l c c}
        \toprule
        Comparison & \textbf{High} & \textbf{Low} \\
        \midrule
        Base vs Few & \(1.47 \times 10^{-2}\) & \(1.39 \times 10^{-8}\) \\
        Base vs P-CoT & \(5.72 \times 10^{-6}\) & \(6.44 \times 10^{-12}\) \\
        Few vs P-CoT & \(1.84 \times 10^{-2}\) & \(2.37 \times 10^{-1}\) \\
        \bottomrule
    \end{tabularx}
    \caption{Mann-Whitney U Test p-values for G2P Conversion task (High/Low)}
    \label{tab:pvalues}
\end{table}

Also, our result is supported by statistically significant p-values from the Mann-Whitney U Test \citep{nachar2008mann}: notably, P-CoT shows significant enhancements over baseline performance for high-frequency (p = \(5.72 \times 10^{-6}\)) and low-frequency words (p = \(6.44 \times 10^{-12}\)). It also surpasses few-shot techniques for high-frequency words (p = \(1.84 \times 10^{-2}\)), though the improvement for low-frequency words is not significant (p = \(2.37 \times 10^{-1}\)). These results highlight P-CoT as an exceptionally effective approach for advancing g2p conversion task performance, effectively adapting to varying word complexities and frequencies.

\subsection{Syllable Counting}
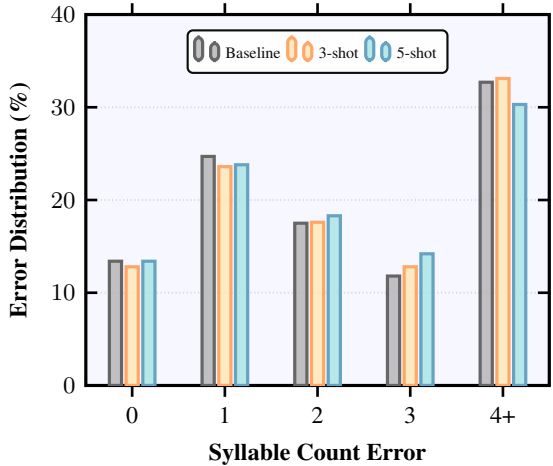
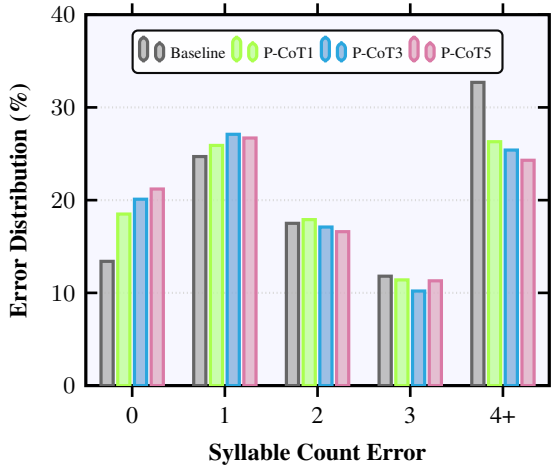
\begin{figure}[!h]
\centering

\begin{subfigure}[b]{0.48\textwidth}
\centering
\begin{tikzpicture}
\begin{axis}[
    width=\textwidth,
    height=6.5cm,
    ybar,
    bar width=0.16cm,
    ymajorgrids=true,
    grid style={densely dotted, gray!30, line width=0.8pt},
    xlabel={Syllable Count Error},
    xlabel style={font=\small\bfseries},
    ylabel={Error Distribution (\%)},
    ylabel style={font=\small\bfseries, align=center},
    title style={font=\large\bfseries, yshift=10pt},
    xticklabels={0, 1, 2, 3, 4+},
    xtick={0,1,2,3,4},
    x tick label style={font=\small},
    xmin=-0.5,
    xmax=4.5,
    ymin=0,
    ymax=40,
    ytick={0,10,20,30,40},
    yticklabel style={
        font=\footnotesize,
        /pgf/number format/fixed,
        /pgf/number format/precision=0
    },
    legend style={
        at={(0.5,0.9)},
        anchor=center,
        legend columns=-1,
        draw=black, 
        line width=0.8pt,
        fill=white,
        font=\tiny,
        inner sep=2pt,
        rounded corners=2pt,
        opacity=0.95,
    },
    axis background/.style={fill=blue!3},
    axis line style={line width=1.2pt, black},
    tick style={line width=1pt, black},
    every axis plot/.append style={thick},
]

\addplot[
    ybar,
    fill=gray!60,
    draw=gray!80!black,
    line width=1.2pt,
    bar shift=-0.22cm,
    fill opacity=0.8,
    draw opacity=1,
] coordinates {
    (0, 13.4)
    (1, 24.7)
    (2, 17.5)
    (3, 11.8)
    (4, 32.7)
};

\addplot[
    ybar,
    fill=orange!65!yellow!30,
    draw=orange!85!red!60,
    line width=1.2pt,
    bar shift=0cm,
    fill opacity=0.8,
    draw opacity=1,
] coordinates {
    (0, 12.8)
    (1, 23.6)
    (2, 17.6)
    (3, 12.8)
    (4, 33.1)
};

\addplot[
    ybar,
    fill=teal!60!cyan!35,
    draw=teal!80!blue!60,
    line width=1.2pt,
    bar shift=0.22cm,
    fill opacity=0.8,
    draw opacity=1,
] coordinates {
    (0, 13.4)
    (1, 23.8)
    (2, 18.3)
    (3, 14.2)
    (4, 30.3)
};

\legend{Baseline, 3-shot, 5-shot}
\end{axis}
\end{tikzpicture}
\caption{Few-shot over Baseline}
\label{fig:syllable_few_shot}
\end{subfigure}
\hfill
\begin{subfigure}[b]{0.48\textwidth}
\centering
\begin{tikzpicture}
\begin{axis}[
    width=\textwidth,
    height=6.5cm,
    ybar,
    bar width=0.16cm,
    ymajorgrids=true,
    grid style={densely dotted, gray!30, line width=0.8pt},
    xlabel={Syllable Count Error},
    xlabel style={font=\small\bfseries},
    ylabel={Error Distribution (\%)},
    ylabel style={font=\small\bfseries, align=center},
    title style={font=\large\bfseries, yshift=10pt},
    xticklabels={0, 1, 2, 3, 4+},
    xtick={0,1,2,3,4},
    x tick label style={font=\small},
    xmin=-0.5,
    xmax=4.5,
    ymin=0,
    ymax=40,
    ytick={0,10,20,30,40},
    yticklabel style={
        font=\footnotesize,
        /pgf/number format/fixed,
        /pgf/number format/precision=0
    },
    legend style={
        at={(0.5,0.9)},
        anchor=center,
        legend columns=-1,
        draw=black, 
        line width=0.8pt,
        fill=white,
        font=\tiny,
        inner sep=2pt,
        rounded corners=2pt,
        opacity=0.95,
    },
    axis background/.style={fill=blue!3},
    axis line style={line width=1.2pt, black},
    tick style={line width=1pt, black},
    every axis plot/.append style={thick},
]

\addplot[
    ybar,
    fill=gray!60,
    draw=gray!80!black,
    line width=1.2pt,
    bar shift=-0.33cm,
    fill opacity=0.8,
    draw opacity=1,
] coordinates {
    (0, 13.4)
    (1, 24.7)
    (2, 17.5)
    (3, 11.8)
    (4, 32.7)
};

\addplot[
    ybar,
    fill=lime!50!green!40,
    draw=lime!70!green!70,
    line width=1.2pt,
    bar shift=-0.11cm,
    fill opacity=0.8,
    draw opacity=1,
] coordinates {
    (0, 18.5)
    (1, 25.9)
    (2, 17.9)
    (3, 11.4)
    (4, 26.3)
};

\addplot[
    ybar,
    fill=cyan!55!blue!45,
    draw=cyan!75!blue!70,
    line width=1.2pt,
    bar shift=0.11cm,
    fill opacity=0.8,
    draw opacity=1,
] coordinates {
    (0, 20.1)
    (1, 27.1)
    (2, 17.1)
    (3, 10.2)
    (4, 25.4)
};

\addplot[
    ybar,
    fill=magenta!50!violet!40,
    draw=magenta!70!violet!70,
    line width=1.2pt,
    bar shift=0.33cm,
    fill opacity=0.8,
    draw opacity=1,
] coordinates {
    (0, 21.2)
    (1, 26.7)
    (2, 16.6)
    (3, 11.3)
    (4, 24.3)
};

\legend{Baseline, P-CoT1, P-CoT3, P-CoT5}

\end{axis}
\end{tikzpicture}
\caption{P-CoT over Baseline}
\label{fig:syllable_pcot}
\end{subfigure}

\caption{Syllable Counting Error Analysis over Baseline (Gray: Baseline, Orange/Teal: Few-shot, Lime/Cyan/Magenta: P-CoT variants)}
\label{fig:syllable_counting}
\end{figure}

The improvement in Table \ref{tab:performance_comparison_syllable} is further illustrated by the error analysis in Figure \ref{fig:syllable_counting}. Our error analysis compares few-shot learning over the baseline across error categories, represented on the x-axis by the number of syllabic errors \([0, 1, 2, 3, 4+]\). Despite using few-shot learning, the distribution remains relatively similar to the baseline, reflecting minimal impact on reducing errors, particularly in the 4+ error  category, showing only a slight decrease with few-shot learning.

Conversely, Figure \ref{fig:syllable_pcot} shows the distributions under P-CoT configurations. Notably, P-CoT prompts lead to a dramatic improvement in reducing instances with 4+ errors, evident from the baseline score of 32.67\% dropping to 24.33\% with P-CoT5. The 4+ error category's significant reduction underscores P-CoT's effectiveness, highlighting its superior ability to guide models towards more accurate syllabic prediction through structured exploration and learning.

These observations strongly support that the P-CoT framework enables LLMs to better interpret syllabic structure, reducing error frequency and distribution markedly. Specifically, the reduction of high-error categories demonstrates that P-CoT facilitates a deeper understanding of syllable structure, allowing models to achieve closer alignment with the correct syllabic count. This pattern affirms that educationally-inspired prompting significantly enhances LLMs' phonological reasoning capabilities, especially for more complex tasks.

\section{Conclusion}

Our study provides a foundational analysis of phonological reasoning within text-based LLMs, a domain that has been largely overlooked in prior research. Contrary to earlier assumptions, our findings reveal a substantial potential for phonological competence within these models. By introducing the Pedagogically-motivated Participatory Chain of Thought (P-CoT) prompting method, grounded in guided discovery learning and scaffolding, we demonstrate significant performance enhancements across rhyme word generation, g2p conversion, and syllable counting tasks, highly surpassing traditional few-shot methods. Our work paves the way for future investigations to refine LLM capabilities in phonology, and encourages the exploration of similar pedagogical approaches in other linguistic domains. Our framework offers a persuasive model for enhancing LLM performance, demonstrating the ability to tap into hidden capabilities using well-structured, education-inspired prompts.

\section{Limitations}

While offering valuable insights into enhancing phonological reasoning within text-based large language models, our study has some limitations. First, the proposed P-CoT prompting framework may not be uniformly optimal across all LLM architectures and sizes; different models could potentially benefit from further tailored prompt designs. Second, the evaluation primarily relies on PhonologyBench, which, while comprehensive, may not capture all nuances of phonological tasks across diverse languages and dialects. Third, our research focused on three specific phonological tasks, and the findings might not generalize across other, untested phonological aspects or linguistic phenomena. Furthermore, the number of example interactions in our prompts do not show a consistently linear relationship with performance improvements. This indicates that simply increasing the number of examples does not always lead to better outcomes, suggesting a potential interplay with factors like example quality, model capacity to process additional context, and inherent task complexity. Lastly, the effectiveness of our method is contingent on the quality and representativeness of the few-shot examples used, which could potentially introduce bias if unaccounted for. Future work should explore cross-linguistic evaluations and adaptive prompt strategies to further refine and validate our approach in broader, more varied contexts.
\section{Ethical Consideration}
Our research adheres to ethical standards, prioritizing transparency, inclusivity, and the responsible deployment of language models. We acknowledge the potential biases inherent in pretrained models and have consciously selected datasets to minimize skew, ensuring diverse representation within the scope of our study. The data and benchmark tasks utilized are publicly available and sourced from reputable repositories to uphold research integrity.

Furthermore, the advancements proposed by our P-CoT prompting are intended to enhance linguistic model capabilities without exacerbating existing issues such as bias or misinformation propagation. While our approach offers significant performance improvements, it is crucial for practitioners to critically evaluate the contexts in which these models are deployed, emphasizing responsible AI usage. This study encourages continuous dialogue and collaboration within the research community to foster ethically sound AI development and application.

\bibliography{references}

\begin{thebibliography}{60}
\providecommand{\natexlab}[1]{#1}

\bibitem[{AI(2024)}]{Ministral8BInstruct2410}
Mistral AI. 2024.
\newblock \href {https://huggingface.co/mistralai/Ministral-8B-Instruct-2410} {Ministral-8b-instruct-2410}.

\bibitem[{Alfieri et~al.(2011)Alfieri, Brooks, Aldrich, and Tenenbaum}]{alfieri2011does}
Louis Alfieri, Patricia~J Brooks, Naomi~J Aldrich, and Harriet~R Tenenbaum. 2011.
\newblock Does discovery-based instruction enhance learning?
\newblock \emph{Journal of educational psychology}, 103(1):1.

\bibitem[{Anand et~al.(2023)Anand, Nussbaum, Duderstadt, Schmidt, and Mulyar}]{gpt4all}
Yuvanesh Anand, Zach Nussbaum, Brandon Duderstadt, Benjamin Schmidt, and Andriy Mulyar. 2023.
\newblock Gpt4all: Training an assistant-style chatbot with large scale data distillation from gpt-3.5-turbo.
\newblock \url{https://github.com/nomic-ai/gpt4all}.

\bibitem[{Anthony(1973)}]{anthony1973learning}
WS~Anthony. 1973.
\newblock Learning to discover rules by discovery.
\newblock \emph{Journal of Educational Psychology}, 64(3):325.

\bibitem[{Anthropic(2024{\natexlab{a}})}]{anthropicClaude3.5Haiku}
Anthropic. 2024{\natexlab{a}}.
\newblock \href {https://www.anthropic.com/claude/haiku} {Claude 3.5 haiku}.

\bibitem[{Anthropic(2024{\natexlab{b}})}]{anthropic2024claude}
AI~Anthropic. 2024{\natexlab{b}}.
\newblock Claude 3.5 sonnet model card addendum.
\newblock \emph{Claude-3.5 Model Card}, 3(6).

\bibitem[{Atox and Clark(2024)}]{atox2024evaluating}
Nathan Atox and Mason Clark. 2024.
\newblock Evaluating large language models through the lens of linguistic proficiency and world knowledge: A comparative study.
\newblock \emph{Authorea Preprints}.

\bibitem[{Baevski et~al.(2020)Baevski, Zhou, Mohamed, and Auli}]{baevski2020wav2vec}
Alexei Baevski, Yuhao Zhou, Abdelrahman Mohamed, and Michael Auli. 2020.
\newblock wav2vec 2.0: A framework for self-supervised learning of speech representations.
\newblock \emph{Advances in neural information processing systems}, 33:12449--12460.

\bibitem[{Brown(1994)}]{brown1994guided}
AL~Brown. 1994.
\newblock Guided discovery in a community of learners.
\newblock \emph{Classroom lessons: Integrating cognitive theory and classroom practice/press/Bradford Books}.

\bibitem[{Brown et~al.(2020)Brown, Mann, Ryder, Subbiah, Kaplan, Dhariwal, Neelakantan, Shyam, Sastry, Askell et~al.}]{brown2020language}
Tom Brown, Benjamin Mann, Nick Ryder, Melanie Subbiah, Jared~D Kaplan, Prafulla Dhariwal, Arvind Neelakantan, Pranav Shyam, Girish Sastry, Amanda Askell, et~al. 2020.
\newblock Language models are few-shot learners.
\newblock \emph{Advances in neural information processing systems}, 33:1877--1901.

\bibitem[{Bruner(1985)}]{bruner1985vygotsky}
Jerome Bruner. 1985.
\newblock Vygotsky: A historical and conceptual perspective.
\newblock \emph{Culture, communication, and cognition: Vygotskian perspectives}, 21:34.

\bibitem[{Bruner(1961)}]{bruner1961act}
Jerome~S Bruner. 1961.
\newblock The act of discovery.
\newblock \emph{Harvard educational review}.

\bibitem[{Bubeck et~al.(2023)Bubeck, Chandrasekaran, Eldan, Gehrke, Horvitz, Kamar, Lee, Lee, Li, Lundberg, Nori, Palangi, Ribeiro, and Zhang}]{bubeck2023sparksartificialgeneralintelligence}
Sébastien Bubeck, Varun Chandrasekaran, Ronen Eldan, Johannes Gehrke, Eric Horvitz, Ece Kamar, Peter Lee, Yin~Tat Lee, Yuanzhi Li, Scott Lundberg, Harsha Nori, Hamid Palangi, Marco~Tulio Ribeiro, and Yi~Zhang. 2023.
\newblock \href {https://arxiv.org/abs/2303.12712} {Sparks of artificial general intelligence: Early experiments with gpt-4}.
\newblock \emph{Preprint}, arXiv:2303.12712.

\bibitem[{Cazden(1979)}]{Cazden1979PeekabooAA}
Courtney~B. Cazden. 1979.
\newblock \href {https://api.semanticscholar.org/CorpusID:140433108} {Peekaboo as an instructional model: Discourse development at home and at school. papers and reports on child language development, no. 17.}

\bibitem[{Chang(2023)}]{chang2023promptinglargelanguagemodels}
Edward~Y. Chang. 2023.
\newblock \href {https://arxiv.org/abs/2303.08769} {Prompting large language models with the socratic method}.
\newblock \emph{Preprint}, arXiv:2303.08769.

\bibitem[{Chen et~al.(2024)Chen, Wang, Xu, Yuan, Zhang, Shi, Xie, Li, Yang, Zhu et~al.}]{chen2024persona}
Jiangjie Chen, Xintao Wang, Rui Xu, Siyu Yuan, Yikai Zhang, Wei Shi, Jian Xie, Shuang Li, Ruihan Yang, Tinghui Zhu, et~al. 2024.
\newblock From persona to personalization: A survey on role-playing language agents.
\newblock \emph{arXiv preprint arXiv:2404.18231}.

\bibitem[{Chowdhery et~al.(2022)Chowdhery, Narang, Devlin, Bosma, Mishra, Roberts, Barham, Chung, Sutton, Gehrmann et~al.}]{chowdhery2022palm}
Aakanksha Chowdhery, Sharan Narang, Jacob Devlin, Maarten Bosma, Gaurav Mishra, Adam Roberts, Paul Barham, Hyung~Won Chung, Charles Sutton, Sebastian Gehrmann, et~al. 2022.
\newblock Palm: Scaling language modeling with pathways.
\newblock \emph{arXiv preprint arXiv:2204.02311}.

\bibitem[{Collins et~al.(1991)Collins, Brown, and Holum}]{Collins1991COGNITIVEAM}
Allan Collins, John~Seely Brown, and Ann Holum. 1991.
\newblock \href {https://api.semanticscholar.org/CorpusID:263308120} {Cognitive apprenticeship: Making thinking visible}.
\newblock \emph{American Educator: The Professional Journal of the American Federation of Teachers}, 15.

\bibitem[{Dao(2023)}]{dao2023flashattention}
Tri Dao. 2023.
\newblock Flashattention-2: Faster attention with better parallelism and work partitioning.
\newblock \emph{arXiv preprint arXiv:2307.08691}.

\bibitem[{Dong et~al.(2022)Dong, Li, Dai, Zheng, Ma, Li, Xia, Xu, Wu, Liu et~al.}]{dong2022survey}
Qingxiu Dong, Lei Li, Damai Dai, Ce~Zheng, Jingyuan Ma, Rui Li, Heming Xia, Jingjing Xu, Zhiyong Wu, Tianyu Liu, et~al. 2022.
\newblock A survey on in-context learning.
\newblock \emph{arXiv preprint arXiv:2301.00234}.

\bibitem[{Dubey et~al.(2024)Dubey, Jauhri, Pandey, Kadian, Al-Dahle, Letman, Mathur, Schelten, Yang, Fan et~al.}]{dubey2024llama}
Abhimanyu Dubey, Abhinav Jauhri, Abhinav Pandey, Abhishek Kadian, Ahmad Al-Dahle, Aiesha Letman, Akhil Mathur, Alan Schelten, Amy Yang, Angela Fan, et~al. 2024.
\newblock The llama 3 herd of models.
\newblock \emph{arXiv preprint arXiv:2407.21783}.

\bibitem[{Dutoit(1997)}]{dutoit1997high}
Thierry Dutoit. 1997.
\newblock High-quality text-to-speech synthesis: An overview.
\newblock \emph{Journal Of Electrical And Electronics Engineering Australia}, 17(1):25--36.

\bibitem[{Ewing~McMahon(2000)}]{ewing2000scaffolding}
Bronwyn~F Ewing~McMahon. 2000.
\newblock Scaffolding: A suitable teaching characteristic in one-to-one teaching in maths recovery.
\newblock In \emph{Mathematics Education Beyond 2000. Twenty Third Annual Conference of the Mathematics Education Research group of Australasia Incorporated. 5-9 July, Fremantle, Western Australia.}, pages 417--423.

\bibitem[{Frantar et~al.(2022)Frantar, Ashkboos, Hoefler, and Alistarh}]{frantar2022gptq}
Elias Frantar, Saleh Ashkboos, Torsten Hoefler, and Dan Alistarh. 2022.
\newblock Gptq: Accurate post-training quantization for generative pre-trained transformers.
\newblock \emph{arXiv preprint arXiv:2210.17323}.

\bibitem[{Guo et~al.(2025)Guo, Yang, Zhang, Song, Zhang, Xu, Zhu, Ma, Wang, Bi et~al.}]{guo2025deepseek}
Daya Guo, Dejian Yang, Haowei Zhang, Junxiao Song, Ruoyu Zhang, Runxin Xu, Qihao Zhu, Shirong Ma, Peiyi Wang, Xiao Bi, et~al. 2025.
\newblock Deepseek-r1: Incentivizing reasoning capability in llms via reinforcement learning.
\newblock \emph{arXiv preprint arXiv:2501.12948}.

\bibitem[{Hardiman et~al.(1986)Hardiman, Pollatsek, and Well}]{hardiman1986learning}
Pamela~Thibodeau Hardiman, Alexander Pollatsek, and Arnold~D Well. 1986.
\newblock Learning to understand the balance beam.
\newblock \emph{Cognition and Instruction}, 3(1):63--86.

\bibitem[{Hu et~al.(2024)Hu, Chen, Yang, Li, Zhang, Chen, and Chng}]{hu2024gentranslatelargelanguagemodels}
Yuchen Hu, Chen Chen, Chao-Han~Huck Yang, Ruizhe Li, Dong Zhang, Zhehuai Chen, and Eng~Siong Chng. 2024.
\newblock \href {https://arxiv.org/abs/2402.06894} {Gentranslate: Large language models are generative multilingual speech and machine translators}.
\newblock \emph{Preprint}, arXiv:2402.06894.

\bibitem[{Jatisunda et~al.(2020)Jatisunda, Suciawati, and Nahdi}]{jatisunda2020discovery}
Mohamad~Gilar Jatisunda, Vici Suciawati, and Dede~Salim Nahdi. 2020.
\newblock Discovery learning with scaffolding to promote mathematical creative thinking ability and self-efficacy.
\newblock \emph{Al-Jabar: Jurnal Pendidikan Matematika}, 11(2):351--370.

\bibitem[{Jiang et~al.(2023)Jiang, Sablayrolles, Mensch, Bamford, Chaplot, Casas, Bressand, Lengyel, Lample, Saulnier et~al.}]{jiang2023mistral}
Albert~Q Jiang, Alexandre Sablayrolles, Arthur Mensch, Chris Bamford, Devendra~Singh Chaplot, Diego de~las Casas, Florian Bressand, Gianna Lengyel, Guillaume Lample, Lucile Saulnier, et~al. 2023.
\newblock Mistral 7b.
\newblock \emph{arXiv preprint arXiv:2310.06825}.

\bibitem[{Jiang et~al.(2024)Jiang, Peng, Feng, Li, and Li}]{jiang2024llms}
Zhuoxuan Jiang, Haoyuan Peng, Shanshan Feng, Fan Li, and Dongsheng Li. 2024.
\newblock Llms can find mathematical reasoning mistakes by pedagogical chain-of-thought.
\newblock \emph{arXiv preprint arXiv:2405.06705}.

\bibitem[{Kirschner et~al.(2006)Kirschner, Sweller, and Clark}]{kirschner2006unguided}
Paul Kirschner, John Sweller, and Richard~E Clark. 2006.
\newblock Why unguided learning does not work: An analysis of the failure of discovery learning, problem-based learning, experiential learning and inquiry-based learning.
\newblock \emph{Educational Psychologist}, 41(2):75--86.

\bibitem[{Liang et~al.(2022)Liang, Bommasani, Lee, Tsipras, Soylu, Yasunaga, Zhang, Narayanan, Wu, Kumar et~al.}]{liang2022holistic}
Percy Liang, Rishi Bommasani, Tony Lee, Dimitris Tsipras, Dilara Soylu, Michihiro Yasunaga, Yian Zhang, Deepak Narayanan, Yuhuai Wu, Ananya Kumar, et~al. 2022.
\newblock Holistic evaluation of language models.
\newblock \emph{arXiv preprint arXiv:2211.09110}.

\bibitem[{Liu et~al.(2023)Liu, Oguz, Zhao, Chang, Stock, Mehdad, Shi, Krishnamoorthi, and Chandra}]{liu2023llm}
Zechun Liu, Barlas Oguz, Changsheng Zhao, Ernie Chang, Pierre Stock, Yashar Mehdad, Yangyang Shi, Raghuraman Krishnamoorthi, and Vikas Chandra. 2023.
\newblock Llm-qat: Data-free quantization aware training for large language models.
\newblock \emph{arXiv preprint arXiv:2305.17888}.

\bibitem[{Loakman et~al.(2023)Loakman, Tang, and Lin}]{loakman2023twistlistresourcesbaselinestongue}
Tyler Loakman, Chen Tang, and Chenghua Lin. 2023.
\newblock \href {https://arxiv.org/abs/2306.03457} {Twistlist: Resources and baselines for tongue twister generation}.
\newblock \emph{Preprint}, arXiv:2306.03457.

\bibitem[{Moreno(2004)}]{moreno2004decreasing}
Roxana Moreno. 2004.
\newblock Decreasing cognitive load for novice students: Effects of explanatory versus corrective feedback in discovery-based multimedia.
\newblock \emph{Instructional science}, 32(1):99--113.

\bibitem[{Nachar et~al.(2008)}]{nachar2008mann}
Nadim Nachar et~al. 2008.
\newblock The mann-whitney u: A test for assessing whether two independent samples come from the same distribution.
\newblock \emph{Tutorials in quantitative Methods for Psychology}, 4(1):13--20.

\bibitem[{OpenAI(2022)}]{openaiGPT3.5turbo}
OpenAI. 2022.
\newblock \href {https://openai.com/index/chatgpt/} {Gpt-3.5 turbo}.

\bibitem[{OpenAI(2023)}]{openai2023gpt}
R~OpenAI. 2023.
\newblock Gpt-4 technical report. arxiv 2303.08774.
\newblock \emph{View in Article}, 2(5).

\bibitem[{Qharabagh et~al.(2024)Qharabagh, Dehghanian, and Rabiee}]{qharabagh2024llmpoweredgraphemetophonemeconversionbenchmark}
Mahta~Fetrat Qharabagh, Zahra Dehghanian, and Hamid~R. Rabiee. 2024.
\newblock \href {https://arxiv.org/abs/2409.08554} {Llm-powered grapheme-to-phoneme conversion: Benchmark and case study}.
\newblock \emph{Preprint}, arXiv:2409.08554.

\bibitem[{Radford et~al.(2023)Radford, Kim, Xu, Brockman, McLeavey, and Sutskever}]{radford2023robust}
Alec Radford, Jong~Wook Kim, Tao Xu, Greg Brockman, Christine McLeavey, and Ilya Sutskever. 2023.
\newblock Robust speech recognition via large-scale weak supervision.
\newblock In \emph{International conference on machine learning}, pages 28492--28518. PMLR.

\bibitem[{Rubenstein et~al.(2023)Rubenstein, Asawaroengchai, Nguyen, Bapna, Borsos, Quitry, Chen, Badawy, Han, Kharitonov et~al.}]{rubenstein2023audiopalm}
Paul~K Rubenstein, Chulayuth Asawaroengchai, Duc~Dung Nguyen, Ankur Bapna, Zal{\'a}n Borsos, F{\'e}lix de~Chaumont Quitry, Peter Chen, Dalia~El Badawy, Wei Han, Eugene Kharitonov, et~al. 2023.
\newblock Audiopalm: A large language model that can speak and listen.
\newblock \emph{arXiv preprint arXiv:2306.12925}.

\bibitem[{Schofield et~al.(2021)Schofield, Wicentowski, and Medero}]{schofield2021learning}
Alexandra Schofield, Richard Wicentowski, and Julie Medero. 2021.
\newblock Learning how to learn nlp: Developing introductory concepts through scaffolded discovery.
\newblock In \emph{Proceedings of the Fifth Workshop on Teaching NLP}, pages 131--137.

\bibitem[{Shanahan(2022)}]{shanahan2022talking}
Murray Shanahan. 2022.
\newblock Talking about large language models.
\newblock \emph{arXiv preprint arXiv:2212.03551}.

\bibitem[{Suvarna et~al.(2024)Suvarna, Khandelwal, and Peng}]{suvarna2024phonologybench}
Ashima Suvarna, Harshita Khandelwal, and Nanyun Peng. 2024.
\newblock Phonologybench: Evaluating phonological skills of large language models.
\newblock \emph{arXiv preprint arXiv:2404.02456}.

\bibitem[{Taylor et~al.(2022)Taylor, Kardas, Cucurull, Scialom, Hartshorn, Saravia, Poulton, Kerkez, and Stojnic}]{taylor2022galactica}
Ross Taylor, Marcin Kardas, Guillem Cucurull, Thomas Scialom, Anthony Hartshorn, Elvis Saravia, Andrew Poulton, Viktor Kerkez, and Robert Stojnic. 2022.
\newblock Galactica: A large language model for science.
\newblock \emph{arXiv preprint arXiv:2211.09085}.

\bibitem[{Team et~al.(2024)Team, Mesnard, Hardin, Dadashi, Bhupatiraju, Pathak, Sifre, Rivi{\`e}re, Kale, Love et~al.}]{team2024gemma}
Gemma Team, Thomas Mesnard, Cassidy Hardin, Robert Dadashi, Surya Bhupatiraju, Shreya Pathak, Laurent Sifre, Morgane Rivi{\`e}re, Mihir~Sanjay Kale, Juliette Love, et~al. 2024.
\newblock Gemma: Open models based on gemini research and technology.
\newblock \emph{arXiv preprint arXiv:2403.08295}.

\bibitem[{Titus(2024)}]{titus2024does}
Lisa~Miracchi Titus. 2024.
\newblock Does chatgpt have semantic understanding? a problem with the statistics-of-occurrence strategy.
\newblock \emph{Cognitive Systems Research}, 83:101174.

\bibitem[{Touvron et~al.(2023)Touvron, Lavril, Izacard, Martinet, Lachaux, Lacroix, Rozi{\`e}re, Goyal, Hambro, Azhar et~al.}]{touvron2023llama}
Hugo Touvron, Thibaut Lavril, Gautier Izacard, Xavier Martinet, Marie-Anne Lachaux, Timoth{\'e}e Lacroix, Baptiste Rozi{\`e}re, Naman Goyal, Eric Hambro, Faisal Azhar, et~al. 2023.
\newblock Llama: Open and efficient foundation language models.
\newblock \emph{arXiv preprint arXiv:2302.13971}.

\bibitem[{Tseng et~al.(2024)Tseng, Huang, Hsiao, Hsu, Foo, Huang, and Chen}]{tseng2024two}
Yu-Min Tseng, Yu-Chao Huang, Teng-Yun Hsiao, Yu-Ching Hsu, Jia-Yin Foo, Chao-Wei Huang, and Yun-Nung Chen. 2024.
\newblock Two tales of persona in llms: A survey of role-playing and personalization.
\newblock \emph{arXiv preprint arXiv:2406.01171}.

\bibitem[{Tuovinen and Sweller(1999)}]{tuovinen1999comparison}
Juhani~E Tuovinen and John Sweller. 1999.
\newblock A comparison of cognitive load associated with discovery learning and worked examples.
\newblock \emph{Journal of educational psychology}, 91(2):334.

\bibitem[{Van~Joolingen et~al.(2005)Van~Joolingen, de~Jong, Lazonder, Savelsbergh, and Manlove}]{van2005co}
Wouter~R Van~Joolingen, Ton de~Jong, Ard~W Lazonder, Elwin~R Savelsbergh, and Sarah Manlove. 2005.
\newblock Co-lab: research and development of an online learning environment for collaborative scientific discovery learning.
\newblock \emph{Computers in human behavior}, 21(4):671--688.

\bibitem[{Vygotsky(1978)}]{vygotsky1978mind}
Lev~S Vygotsky. 1978.
\newblock \emph{Mind in society: The development of higher psychological processes}, volume~86.
\newblock Harvard university press.

\bibitem[{Wang et~al.(2023)Wang, Mao, Wu, Ge, Wei, and Ji}]{wang2023unleashing}
Zhenhailong Wang, Shaoguang Mao, Wenshan Wu, Tao Ge, Furu Wei, and Heng Ji. 2023.
\newblock Unleashing the emergent cognitive synergy in large language models: A task-solving agent through multi-persona self-collaboration.
\newblock \emph{arXiv preprint arXiv:2307.05300}.

\bibitem[{Wei et~al.(2022)Wei, Wang, Schuurmans, Bosma, Xia, Chi, Le, Zhou et~al.}]{wei2022chain}
Jason Wei, Xuezhi Wang, Dale Schuurmans, Maarten Bosma, Fei Xia, Ed~Chi, Quoc~V Le, Denny Zhou, et~al. 2022.
\newblock Chain-of-thought prompting elicits reasoning in large language models.
\newblock \emph{Advances in neural information processing systems}, 35:24824--24837.

\bibitem[{Wood et~al.(1976)Wood, Bruner, and Ross}]{wood1976role}
David Wood, Jerome~S Bruner, and Gail Ross. 1976.
\newblock The role of tutoring in problem solving.
\newblock \emph{Journal of child psychology and psychiatry}, 17(2):89--100.

\bibitem[{Xue et~al.(2021)Xue, Song, Wu, Tan, Zhang, Qin, Zhang, and Liu}]{xue2021deeprapper}
Lanqing Xue, Kaitao Song, Duocai Wu, Xu~Tan, Nevin~L Zhang, Tao Qin, Wei-Qiang Zhang, and Tie-Yan Liu. 2021.
\newblock Deeprapper: Neural rap generation with rhyme and rhythm modeling.
\newblock \emph{arXiv preprint arXiv:2107.01875}.

\bibitem[{Yang et~al.(2024)Yang, Yang, Hui, Zheng, Yu, Zhou, Li, Li, Liu, Huang et~al.}]{yang2024qwen2}
An~Yang, Baosong Yang, Binyuan Hui, Bo~Zheng, Bowen Yu, Chang Zhou, Chengpeng Li, Chengyuan Li, Dayiheng Liu, Fei Huang, et~al. 2024.
\newblock Qwen2 technical report.
\newblock \emph{arXiv preprint arXiv:2407.10671}.

\bibitem[{Yu et~al.(2024)Yu, Liu, Hou, Ding, Tao, and Zhang}]{yu2024self}
Tengfei Yu, Xuebo Liu, Zhiyi Hou, Liang Ding, Dacheng Tao, and Min Zhang. 2024.
\newblock Self-powered llm modality expansion for large speech-text models.
\newblock In \emph{Proceedings of the 2024 Conference on Empirical Methods in Natural Language Processing}, pages 12401--12417.

\bibitem[{Zhang et~al.(2023{\natexlab{a}})Zhang, Li, Zhang, Zhan, Wang, Zhou, and Qiu}]{zhang2023speechgpt}
Dong Zhang, Shimin Li, Xin Zhang, Jun Zhan, Pengyu Wang, Yaqian Zhou, and Xipeng Qiu. 2023{\natexlab{a}}.
\newblock Speechgpt: Empowering large language models with intrinsic cross-modal conversational abilities.
\newblock \emph{arXiv preprint arXiv:2305.11000}.

\bibitem[{Zhang et~al.(2023{\natexlab{b}})Zhang, Zhang, Song, Zhang, Wu, and He}]{zhang2023prosody}
Ya-Jie Zhang, Chao Zhang, Wei Song, Zhengchen Zhang, Youzheng Wu, and Xiaodong He. 2023{\natexlab{b}}.
\newblock Prosody modelling with pre-trained cross-utterance representations for improved speech synthesis.
\newblock \emph{IEEE/ACM Transactions on Audio, Speech, and Language Processing}, 31:2812--2823.

\end{thebibliography}

\appendix

\section{Details on P-CoT Prompting Method}\label{sec:appendix}
We explained that our approach delivers temporary, reciprocal guidance through targeted hints, clear definitions of key concepts, and the assignment of specific tasks. This design also enables models to reach a deeper level of understanding by emphasizing the learner’s active engagement. In developing our detailed prompts, we specifically consulted real-world educational applications of these methods, as documented by 
\citet{wood1976role}, \citet{Collins1991COGNITIVEAM}, \citet{ewing2000scaffolding}, \citet{jatisunda2020discovery}, and \citet{schofield2021learning}. In this section, we provide a detailed overview of how we construct prompts based on the educational principles, illustrating each component of our framework.

\citet{Collins1991COGNITIVEAM} propose a reciprocal teaching approach in which teachers and students alternate roles. Building on this idea, we tested two scenarios: one where the model served as the teacher and another where it acted as the student. Here we present the scenario that achieved the best performance for each task. The model functioned optimally as the teacher for syllable counting and rhyme word generation, while it achieved higher accuracy as the student in g2p conversion tasks.
Although we report the optimal role for each task, the effectiveness may stem more from exposing the model to an interactive learning context that offers cues for reasoning.

\subsection{Rhyme Word Generation}
\label{appendix:rhyme-table 9}
Here is a detailed overview of our 3-shot P-CoT prompt for the rhyme word generation task. As shown in Table~\ref{tab:3Peda-CoT-part1}, we adopt a \textbf{role-based prompt} format, implemented using a system/user message structure common in instruction-tuned LLMs. In this setting, the role-setting message assigns the model a teacher persona by stating:

\begin{tcolorbox}[colback=blue!5!white,colframe=blue!50!black,title= Role Assignment,fonttitle=\bfseries\small]
\textit{"You are an expert in American English phonology, phonetics, and morphology."} \\
\textit{"Your goal is to assist the student in discovering words that rhyme with a given word using discovery learning methods."}
\end{tcolorbox}

Following \citet{jatisunda2020discovery}, we also provide a clear definition of the core concept to prevent misconceptions:

\begin{tcolorbox}[colback=gray!8!white,colframe=gray!60!black,leftrule=3pt,rightrule=0pt,toprule=0pt,bottomrule=0pt]
\textbf{Conceptual Foundation:} \textit{"Rhyming words are words that have the same ending sound. In simpler terms, it is the repetition of similar ending sounds."}
\end{tcolorbox}

Before the guided exploration begins, the \textbf{User} adopts the student role, declaring:

\begin{tcolorbox}[colback=green!5!white,colframe=green!50!black,arc=2pt,boxrule=0.8pt]
\textbf{Student Readiness:} \textit{"I'm ready to learn about rhyming words and how to identify them."}
\end{tcolorbox}

In line with \citet{jatisunda2020discovery} and \citet{wood1976role}, this statement frames the user as an attentive, motivated learner. The teacher (i.e., the \textbf{Assistant}) then reaffirms essential information:

\begin{tcolorbox}[colback=orange!8!white,colframe=orange!60!black,arc=1.5pt,boxrule=1pt]
\textbf{Scaffolding 1(Concept Understanding):} \textit{"Let's start by understanding what rhyming words are. Remember, they share the same ending sounds."}
\end{tcolorbox}

Rather than providing the answer outright, the teacher encourages the student to identify the ending sound of a sample word. This scaffolding is repeated three times in the 3-shot P-CoT, allowing the student to actively engage in the discovery process.
When the student appears unsure, the teacher further breaks down the task:

\begin{tcolorbox}[colback=purple!6!white,colframe=purple!50!black,boxrule=1pt,arc=3pt,left=8pt,right=8pt,top=6pt,bottom=6pt]
\textbf{Scaffolding 2 (Task Decomposition):} \\
\textit{"Start by identifying the ending sound of 'information' and try to think of other words with a similar ending."}
\end{tcolorbox}

This stepwise approach divides the problem into smaller tasks, promoting deeper understanding \citep{Collins1991COGNITIVEAM, ewing2000scaffolding}. By embedding these teacher–student interactions directly into the prompt, our 3-shot P-CoT design demonstrates how guided discovery learning can ultimately enhance performance in rhyme word generation.

\subsection{G2P Conversion}
\label{appendix:g2p-table 8}
This is a detailed overview of our 3-shot P-CoT prompt for g2p conversion, as illustrated in Table~\ref{tab:3Peda-CoT-part1}. We adopt a \textbf{role-based prompt} format, but in this scenario, we do not explicitly assign a student persona to the model. Instead, the role-setting message states:

\begin{tcolorbox}[colback=blue!5!white,colframe=blue!50!black,title= Role Assignment,fonttitle=\bfseries\small]
\textit{"You are a leading expert in General American English (GAE) phonology. Your specialty is converting written words (graphemes) into their precise GAE pronunciation (phonemes)."}
\end{tcolorbox}

This setup provides the model with a phonology-focused persona, implicitly allowing the \textbf{Assistant} to fulfill the role of an attentive learner, consistent with \citet{jatisunda2020discovery} and \citet{wood1976role}.

The teacher (i.e., the \textbf{User}) initiates the learning process by offering an example, saying:

\begin{tcolorbox}[colback=green!6!white,colframe=green!50!black,arc=2pt,boxrule=0.8pt]
\textbf{Scaffolding 1 (Provide hints):} \textit{"Let's analyze the word 'apparently' using standard GAE pronunciation. Show me its exact phonemic transcription, paying special attention to American rhotic sounds and stress patterns."}
\end{tcolorbox}

Rather than supplying a direct solution, this prompt aligns with discovery learning by prompting the student to explore the g2p conversion process. Hints about GAE-specific pronunciations (e.g., rhotic sounds) serve as scaffolding, helping the student avoid misconceptions and focus on critical aspects of phoneme conversion.

In response, the student (i.e., the \textbf{Assistant}) details its reasoning, for instance:

\begin{tcolorbox}[colback=orange!8!white,colframe=orange!60!black,arc=1.5pt,boxrule=1pt]
\textbf{Student Phonological Analysis:} \textit{"In GAE, we have the unstressed initial \textipa{@p}, followed by the stressed syllable \textipa{"Er} with the characteristic American rhotic \textipa{r}, then the reduced \textipa{@nt}, and finally \textipa{li}."}
\end{tcolorbox}

By explicitly describing its thought process, the student demonstrates the inquiry-driven nature of discovery learning \citep{bruner1961act}, gradually refining its approach through teacher-provided scaffolding \citep{wood1976role}. This interaction repeats three times in the 3-shot P-CoT, each iteration offering guidance on tricky GAE features.

Finally, recognizing that scaffolding should be gradually removed, the last request no longer provides word-specific hints. Instead, the teacher offers general guidelines and then states:

\begin{tcolorbox}[colback=purple!6!white,colframe=purple!50!black,boxrule=1pt,arc=3pt,left=8pt,right=8pt,top=6pt,bottom=6pt]
\textbf{Independent Application Task:} \\
\textit{"Now, provide the precise General American English phonemic transcription for '\{text\}'."}
\end{tcolorbox}

This gradual removal of support requires the student to apply what it has learned independently, aligning with the principle that scaffolding is intended to be temporary. Consequently, the student takes an active role in completing the g2p conversion, illustrating how our 3-shot P-CoT framework effectively integrates discovery learning and scaffolding to enhance phonological reasoning.

\subsection{Syllable Counting}
\label{appendix:syllable-table 8} 
Below is a detailed description of our 3-shot P-CoT prompt for the syllable counting task, illustrating its key components and flow. As with the other tasks, we adopt a \textbf{role-based prompt} format. As shown in Table~\ref{tab:3Peda-CoT-part2}, we begin by assigning the model a teacher persona through the role-setting message, which states:

\begin{tcolorbox}[colback=blue!5!white,colframe=blue!50!black,title= Role Assignment,fonttitle=\bfseries\small]
\textit{"You are an expert in American English phonology and phonetics."} \\
\textit{"Your goal is to assist the student in counting the total number of syllables in a given sentence using discovery learning methods."}
\end{tcolorbox}

Before the guided learning begins, the \textbf{User} adopts the student role by declaring:

\begin{tcolorbox}[colback=green!5!white,colframe=green!50!black,arc=2pt,boxrule=0.8pt]
\textbf{Student Readiness:} \textit{"I'm ready to learn about syllables and how to count them in a sentence."}
\end{tcolorbox}

Following \citet{jatisunda2020discovery} and \citet{wood1976role}, this statement establishes an attentive, motivated learner.
The teacher (i.e., the \textbf{Assistant}) then reiterates essential information to prevent misconceptions, saying:

\begin{tcolorbox}[colback=orange!8!white,colframe=orange!60!black,arc=1.5pt,boxrule=1pt]
\textbf{Scaffolding 1(Concept Understanding):} \textit{"Let's start by understanding what a syllable is. Remember, a syllable has one vowel sound."}
\end{tcolorbox}

The teacher first offers an example and encourages the student to attempt a solution, reflecting the discovery learning principle that learners must actively engage in problem-solving.
If the student appears uncertain, the teacher further breaks down the task, for instance as suggested in \citet{Collins1991COGNITIVEAM} and \citet{ewing2000scaffolding}:

\begin{tcolorbox}[colback=purple!6!white,colframe=purple!50!black,boxrule=1pt,arc=3pt,left=8pt,right=8pt,top=6pt,bottom=6pt]
\textbf{Scaffolding 2 (Task Decomposition):} \\
\textit{"Start by identifying the vowel sounds in each word."}
\end{tcolorbox}

This progressive guidance fosters discovery by dividing the overall process into smaller steps. When the student completes each step (e.g., counting the vowel sounds per word), the teacher offers additional support:

\begin{tcolorbox}[colback=green!5!white,colframe=green!50!black,arc=2pt,boxrule=0.8pt]
\textbf{Scaffolding 3 (Task Decomposition):} \textit{"Exactly! Now, let's sum them all up."}
\end{tcolorbox}

In the 3-shot P-CoT setup, this cycle repeats three times. Finally, to maintain the principle that scaffolding is temporary, the final request does not break the task into sub-steps. Instead, it offers only a hint about identifying each vowel sound before instructing the teacher-model to:

\begin{tcolorbox}[colback=red!6!white,colframe=red!50!black,boxrule=1pt,arc=2pt,left=6pt,right=6pt,top=5pt,bottom=5pt]
\textbf{Independent Application Task:} \textit{"Count the total number of syllables in the given word (in Target Sentence)."}
\end{tcolorbox}

Through this reciprocal interplay, guided discovery learning is effectively realized, demonstrating how our P-CoT prompt facilitates deeper understanding in the syllable counting task.

\begin{figure*}[t]
\centering
\resizebox{\textwidth}{!}{%
\begin{tikzpicture}[
    system/.style={fill=gray!8, draw=gray!40, line width=0.8pt, rounded corners=2pt, inner sep=6pt, text width=12cm},
    user/.style={fill=blue!6, draw=blue!30, line width=0.8pt, rounded corners=2pt, inner sep=5pt, text width=8cm},
    assistant/.style={fill=orange!6, draw=orange!30, line width=0.8pt, rounded corners=2pt, inner sep=5pt, text width=8cm},
    example/.style={fill=black!3, draw=black!20, line width=0.6pt, rounded corners=1pt, inner sep=5pt, text width=12cm},
    task/.style={fill=red!6, draw=red!30, line width=0.8pt, rounded corners=2pt, inner sep=6pt, text width=12cm},
]

\draw[gray!25, line width=2pt, densely dotted] (-5,-2.5) -- (-5,-3.5) -- (5,-3.5) -- (5,-5.5) -- (-5,-5.5) -- (-5,-6.5) -- (5,-6.5) -- (5,-7.5) -- (-5,-7.5) -- (-5,-8.5) -- (5,-8.5) -- (5,-9.5) -- (-5,-9.5) -- (-5,-11.5) -- (5,-11.5) -- (5,-12.5) -- (-5,-12.5) -- (-5,-13.5) -- (5,-13.5) -- (5,-14.5) -- (-5,-14.5) -- (-5,-15.5) -- (5,-15.5) -- (0,-16.5) -- (0,-17.5);

\draw[black, line width=2pt, rounded corners=3pt] (-7,0.5) rectangle (7,-19.2);

\node[anchor=center, font=\large\bfseries\sffamily] at (0,0) {
    P-CoT5 Prompting Flow (Syllable Counting)
};

\node[system, anchor=north, font=\tiny] at (0,-1) {
    \textbf{System:} You are an expert in American English phonology and phonetics. A syllable is a unit of pronunciation having exactly one vowel sound, forming the whole or a part of a word. Your goal is to assist the student in counting the total number of syllables in a given sentence using discovery learning methods.
};

\node[user, anchor=north west, font=\tiny] at (-6.5,-2.5) {
    \textbf{User:} "I'm ready to learn about syllables and how to count them in a sentence."
};

\node[assistant, anchor=north east, font=\tiny] at (6.5,-3.5) {
    \textbf{Assistant:} "Great! Let's start by understanding what a syllable is. Remember, a syllable has one vowel sound. Now let's try an example together."
};

\node[example, anchor=north, font=\tiny] at (0,-4.5) {
    \textbf{Example 1:} "Sentence: 'Grace has resigned herself to simply completing the upbringing of her teenage daughter.' How many syllables do you think it has?"
};

\node[assistant, anchor=north east, font=\tiny] at (6.5,-5.5) {
    \textbf{Assistant:} "Let's break it down together. Start by identifying the vowel sounds in each word."
};

\node[user, anchor=north west, font=\tiny] at (-6.5,-6.5) {
    \textbf{User:} "Grace (1), has (1), resigned (2), herself (2), to (1), simply (2), completing (3), the (1), upbringing (3), of (1), her (1), teenage (2), daughter (2)"
};

\node[assistant, anchor=north east, font=\tiny] at (6.5,-7.5) {
    \textbf{Assistant:} "Exactly! Now, let's sum them all up."
};

\node[user, anchor=north west, font=\tiny] at (-6.5,-8.5) {
    \textbf{User:} "22"
};

\node[assistant, anchor=north east, font=\tiny] at (6.5,-9.5) {
    \textbf{Assistant:} "That's correct! There are 22 syllables in that sentence."
};

\node[example, anchor=north, font=\tiny] at (0,-10.5) {
    \textbf{Example 2:} "Sentence: 'This story is about a young girl's redemption in a small town.' How many syllables do you think it has?"
};

\node[assistant, anchor=north east, font=\tiny] at (6.5,-11.5) {
    \textbf{Assistant:} "Let's work through this one as well. Identify the vowel sounds in each word."
};

\node[user, anchor=north west, font=\tiny] at (-6.5,-12.5) {
    \textbf{User:} "This (1), story (2), is (1), about (2), a (1), young (1), girl's (1), redemption (3), in (1), a (1), small (1), town (1)"
};

\node[assistant, anchor=north east, font=\tiny] at (6.5,-13.5) {
    \textbf{Assistant:} "Good job! Now add them together."
};

\node[user, anchor=north west, font=\tiny] at (-6.5,-14.5) {
    \textbf{User:} "16"
};

\node[assistant, anchor=north east, font=\tiny] at (6.5,-15.5) {
    \textbf{Assistant:} "Correct again! It has 16 syllables."
};

\node[example, anchor=north, font=\tiny] at (0,-16.5) {
    \textbf{Examples 3-5:} Similar pattern with sentences yielding 13, 16, and 20 syllables respectively, following identical discovery-learning methodology.
};

\node[task, anchor=north, font=\tiny] at (0,-17.5) {
    \textbf{Final Task:} "Now it's your turn! Count the \textit{total} number of syllables in the given word (in Target Sentence): '\{text\}.' Do not randomly generate the number; rather, you must analyze the vowel sounds and sum them accurately. Provide \textit{only} the total count."
};

\node[anchor=west, font=\tiny\sffamily, text=gray!60] at (-6.8,-1) {Dialogue Flow};
\node[anchor=east, font=\tiny\sffamily, text=gray!60] at (6.8,-1) {Discovery Learning};

\end{tikzpicture}
}
\caption{
\textbf{P-CoT5 Prompting Flow for Syllable Counting Task.}}
\label{fig:pcot5_syllable_counting}
\end{figure*}

\clearpage

\begin{table*}[t]
\centering
\footnotesize
\begin{tabularx}{\textwidth}{p{3cm} X}
\toprule
\textbf{Task} & \textbf{Prompt} \\
\midrule
\textbf{P-CoT1: Rhyme Word Generation} 
& 
\textbf{System:} 
You are an expert in American English phonology, phonetics, and morphology. Rhyming words share the same ending sound. As an expert, guide the user to discover 5 different words that rhyme with a given word by giving subtle hints and ensuring they find the exact answer.

\textbf{User:} 
“As the Prominent Phonology Professor, I know that rhyming words have the same ending sound. 
Give exactly 5 \emph{different} words that rhyme with ‘\{text\}’, separated by commas. 
Do not include any breakdown or additional explanation.”

\\
\midrule
\textbf{P-CoT1: G2P Conversion} 
& 
\textbf{System:} 
You are a leading expert in General American English (GAE) phonology. Your specialty is converting written words (graphemes) into their precise GAE pronunciation (phonemes). 

\textbf{User (example 1):} 
“Let's analyze the word ‘apparently’ using standard GAE pronunciation. Show me its exact phonemic transcription, paying special attention to American rhotic sounds and stress patterns.”

\textbf{Assistant:} 
“The General American English pronunciation of ‘apparently’ is \textipa{/@p"E\*r@ntli/}. 
In GAE, we have the unstressed initial \textipa{@p}, followed by the stressed syllable 
\textipa{"E\*r} with the characteristic American rhotic \textipa{\*r}, then 
the reduced \textipa{@nt}, and finally \textipa{li}.”

\textbf{User (final request):} 
“Now, provide the precise General American English phonemic transcription for ‘\{text\}’. 
Focus on accurate vowels, rhotics (\textipa{\*r}), and stress. 
Give me \emph{only} its complete GAE phonemic transcription.”

\\
\midrule
\textbf{P-CoT1: Syllable Counting} 
& 
\textbf{System:} 
You are an expert in American English phonology and phonetics. A syllable is a unit of pronunciation having exactly one vowel sound. Your goal is to assist in counting the total number of syllables in a given sentence using discovery learning.

\textbf{User:} 
“I'm ready to learn about syllables and how to count them in a sentence.”

\textbf{Assistant:} 
“Great! Let's start by understanding what a syllable is. Remember, a syllable has one vowel sound. Now let's try an example together.”

\textbf{User (example 1):} 
“Sentence: ‘Grace has resigned herself to simply completing the upbringing of her teenage daughter.’ How many syllables do you think it has?”

\textbf{Assistant:} 
“Let's break it down together. Identify the vowel sounds in each word.”

\textbf{User:} 
“Grace (1), has (1), resigned (2) ... teenage (2), daughter (2)”

\textbf{Assistant:} 
“Exactly! Now, let's sum them all up.”

\textbf{User:} 
“22”

\textbf{Assistant:} 
“That's correct! There are 22 syllables in that sentence.”

\textbf{User (final request):} 
“Now it's your turn! Count the TOTAL number of syllables in the given word (in Target Sentence): ‘\{text\}’. 
Do not randomly guess the number; you must identify vowel sounds and add them up. Provide \emph{only} the total count.”

\\
\bottomrule
\end{tabularx}
\caption{P-CoT1 tasks (Rhyme Word Generation, G2P Conversion, Syllable Counting) with sample prompts.}
\label{tab:CoT-prompts}
\end{table*}

\begin{table*}[t]
\centering
\footnotesize
\begin{tabularx}{\textwidth}{p{3cm} X}
\toprule
\textbf{Task} & \textbf{Prompt} \\
\midrule

\textbf{P-CoT3: Rhyme Word Generation}
&
\textbf{System:}
You are an expert in American English phonology, phonetics, and morphology. 
Rhyming words are words that have the same ending sound. 
In simpler terms, it is the repetition of similar ending sounds. 
Your goal is to assist the student in discovering words that rhyme with a given word using discovery learning methods.

\textbf{User:}
“I'm ready to learn about rhyming words and how to identify them.”

\textbf{Assistant:}
“Great! Let's start by understanding what rhyming words are. 
Remember, they share the same ending sounds. 
Now let's try an example together.”

\textbf{User (example 1):}
“Word: ‘information.’ 
Can you think of any words that rhyme with ‘information’?”

\textbf{Assistant:}
“Let's break it down together. 
Start by identifying the ending sound of ‘information’ and try to think of other words with a similar ending.”

\textbf{User:}
“Here are some words that rhyme with ‘information’: isolation, operation, conversation, corporation, demonstration.”

\textbf{Assistant:}
“That's correct! Those are great examples of rhyming words with ‘information’.”

\textbf{User (example 2):}
“Word: ‘available.’ 
Can you think of any words that rhyme with ‘available’?”

\textbf{Assistant:}
“Let's work through this one as well. 
Identify the ending sound of ‘available’ and find other words with similar endings.”

\textbf{User:}
“Here are some words that rhyme with ‘available’: distrainable, explainable, restrainable, retainable, retrainable.”

\textbf{Assistant:}
“Correct again! Those are good examples of rhyming words with ‘available’.”

\textbf{User (example 3):}
“Word: ‘transport.’ 
Can you think of any words that rhyme with ‘transport’?”

\textbf{Assistant:}
“Let's do the same process and find words with similar ending sounds to ‘transport’.”

\textbf{User:}
“Here are some words that rhyme with ‘transport’: passport, escort, report, resort, retort.”

\textbf{Assistant:}
“Perfect! Those words rhyme with ‘transport’.”

\textbf{User (final request):}
“Now it's your turn! Identify the ending sound and find words that rhyme with the given word. 
Word: ‘\{text\}.’ 
Give exactly 5 \emph{different} words that rhyme with it, separated by commas. 
Begin with the phrase ‘Here are some words that rhyme with \{text\}:’”
\\
\midrule

\textbf{P-CoT3: G2P Conversion} 
& 
\textbf{System:}
You are a leading expert in General American English (GAE) phonology. 
Your specialty is converting written words (graphemes) into their precise GAE pronunciation (phonemes).

\textbf{User (example 1):}
“Let's analyze the word ‘apparently’ using standard GAE pronunciation. 
Show me its exact phonemic transcription, paying special attention to American rhotic sounds and stress patterns.”

\textbf{Assistant:}
“The General American English pronunciation of ‘apparently’ is 
\textipa{/@p"Er@ntli/}. 
In GAE, we have the unstressed initial \textipa{@p}, 
followed by the stressed syllable \textipa{"Er} with the characteristic American rhotic \textipa{r}, 
then the reduced \textipa{@nt}, and finally \textipa{li}.”

\textbf{User (example 2):}
“For our next GAE analysis, let's examine the word ‘calorie.’ 
What is its precise General American English phonemic transcription? 
Remember to account for the distinctive r-colored vowel that characterizes GAE.”

\textbf{Assistant:}
“The General American English pronunciation of ‘calorie’ is 
\textipa{/"k{\ae}l\textrhookschwa i/}. 
In GAE, we have the stressed syllable \textipa{"k{\ae}l}, 
followed by the r-colored schwa \textipa{\textrhookschwa}, 
and ending with \textipa{i}.”

\textbf{User (example 3):}
“Let's examine ‘freshman’ in General American English. 
What is its precise GAE phonemic transcription? 
Note particularly how the rhotic sound manifests in American pronunciation.”

\textbf{Assistant:}
“The General American English pronunciation of ‘freshman’ is 
\textipa{/"f\*rESm@n/}. 
We begin with the consonant cluster \textipa{f\*r} featuring the American rhotic, 
followed by the stressed \textipa{"E S}, and ending with the reduced syllable \textipa{m@n}.”

\textbf{User (final request):}
“Now, provide the precise General American English phonemic transcription for ‘\{text\}.’ 
Focus on accurate representation of American vowels, rhotics (\textipa{r}), and stress patterns. 
Give me \emph{only} its complete GAE phonemic transcription.”
\\

\bottomrule
\end{tabularx}
\caption{P-CoT3 tasks (Rhyme Word Generation and G2P Conversion) with sample prompts.}
\label{tab:3Peda-CoT-part1}
\end{table*}

\begin{table*}[t]
\centering
\footnotesize
\begin{tabularx}{\textwidth}{p{3cm} X}
\toprule
\textbf{Task} & \textbf{Prompt} \\
\midrule

\textbf{P-CoT3: Syllable Counting}
&
\textbf{System:}
You are an expert in American English phonology and phonetics. 
A syllable is a unit of pronunciation having exactly one vowel sound. 
Your goal is to assist the student in counting the total number of syllables in a given sentence using discovery learning methods.

\textbf{User:}
“I'm ready to learn about syllables and how to count them in a sentence.”

\textbf{Assistant:}
“Great! Let's start by understanding what a syllable is. 
Remember, a syllable has one vowel sound. 
Now let's try an example together.”

\textbf{User (example 1):}
“Sentence: ‘Grace has resigned herself to simply completing the upbringing of her teenage daughter.’ 
How many syllables do you think it has?”

\textbf{Assistant:}
“Let's break it down together. 
Start by identifying the vowel sounds in each word.”

\textbf{User:}
“Grace (1), has (1), resigned (2), herself (2), to (1), simply (2), completing (3), the (1), upbringing (3), of (1), her (1), teenage (2), daughter (2)”

\textbf{Assistant:}
“Exactly! Now, let's sum them all up.”

\textbf{User:}
“22”

\textbf{Assistant:}
“That's correct! There are 22 syllables in that sentence.”

\textbf{User (example 2):}
“Sentence: ‘This story is about a young girl's redemption in a small town.’ 
How many syllables do you think it has?”

\textbf{Assistant:}
“Let's work through this one as well. 
Identify the vowel sounds in each word.”

\textbf{User:}
“This (1), story (2), is (1), about (2), a (1), young (1), girl's (1), redemption (3), in (1), a (1), small (1), town (1)”

\textbf{Assistant:}
“Good job! Now add them together.”

\textbf{User:}
“16”

\textbf{Assistant:}
“Correct again! It has 16 syllables.”

\textbf{User (example 3):}
“Sentence: ‘The one thing that hasn't happened is a proposal.’ 
How many syllables do you think it has?”

\textbf{Assistant:}
“Let's do the same process and find the vowel sounds in each word.”

\textbf{User:}
“The (1), one (1), thing (1), that (1), hasn’t (2), happened (2), is (1), a (1), proposal (3)”

\textbf{Assistant:}
“Perfect! Now, sum them up.”

\textbf{User:}
“13”

\textbf{Assistant:}
“Well done! The total is 13 syllables.”

\textbf{User (final request):}
“Now it's your turn! Count the \emph{total} number of syllables in the given word (in Target Sentence): 
‘\{text\}.’ 
Do not randomly generate the number; rather, identify each vowel sound and add them up. 
Provide \emph{only} the total count.”
\\

\bottomrule
\end{tabularx}
\caption{P-CoT3 task (Syllable Counting) with sample prompts.}
\label{tab:3Peda-CoT-part2}
\end{table*}

\begin{table*}[t]
\centering
\footnotesize
\begin{tabularx}{\textwidth}{p{3cm} X}
\toprule
\textbf{Task} & \textbf{Prompt} \\
\midrule

\textbf{P-CoT5: Rhyme Word Generation}
&
\textbf{System:}
You are an expert in American English phonology, phonetics, and morphology. 
Rhyming words are words that have the same ending sound. 
In simpler terms, it can be defined as the repetition of similar ending sounds. 
Your goal is to assist the student in discovering words that rhyme with a given word using discovery learning methods.

\textbf{User:}
“I'm ready to learn about rhyming words and how to identify them.”

\textbf{Assistant:}
“Great! Let's start by understanding what rhyming words are. 
Remember, they share the same ending sounds. 
Now let's try an example together.”

\textbf{User (example 1):}
“Word: ‘information.’ 
Can you think of any words that rhyme with ‘information’?”

\textbf{Assistant:}
“Let's break it down together. 
Start by identifying the ending sound of ‘information’ and try to think of other words with a similar ending.”

\textbf{User:}
“Here are some words that rhyme with ‘information’: isolation, operation, conversation, corporation, demonstration.”

\textbf{Assistant:}
“That's correct! Those are great examples of rhyming words with ‘information’.”

\textbf{User (example 2):}
“Word: ‘available.’ 
Can you think of any words that rhyme with ‘available’?”

\textbf{Assistant:}
“Let's work through this one as well. 
Identify the ending sound of ‘available’ and find other words with similar endings.”

\textbf{User:}
“Here are some words that rhyme with ‘available’: distrainable, explainable, restrainable, retainable, retrainable.”

\textbf{Assistant:}
“Correct again! Those are good examples of rhyming words with ‘available’.”

\textbf{User (example 3):}
“Word: ‘transport.’ 
Can you think of any words that rhyme with ‘transport’?”

\textbf{Assistant:}
“Let's do the same process and find words with similar ending sounds to ‘transport’.”

\textbf{User:}
“Here are some words that rhyme with ‘transport’: passport, escort, report, resort, retort.”

\textbf{Assistant:}
“Perfect! Those words rhyme with ‘transport’.”

\textbf{User (example 4):}
“Word: ‘interesting.’ 
Can you think of any words that rhyme with ‘interesting’?”

\textbf{Assistant:}
“Let's find words that have a similar ending sound to ‘interesting’.”

\textbf{User:}
“Here are some words that rhyme with ‘interesting’: beginning, interrupting, diminishing, investing, referencing.”

\textbf{Assistant:}
“Well done! Those are excellent examples of rhyming words with ‘interesting’.”

\textbf{User (example 5):}
“Word: ‘technology.’ 
Can you think of any words that rhyme with ‘technology’?”

\textbf{Assistant:}
“Try to recognize the ending sound of ‘technology’ and think of other words that share this ending.”

\textbf{User:}
“Here are some words that rhyme with ‘technology’: eternity, innocuity, unity, activity, amusingly.”

\textbf{Assistant:}
“Excellent! Those words rhyme with ‘technology’.”

\textbf{User (final request):}
“Now it's your turn! Identify the ending sound and find words that rhyme with the given word. 
Word: ‘\{text\}.’ 
Give exactly 5 \emph{different} words that rhyme with it, separated by commas. 
Begin with the phrase ‘Here are some words that rhyme with \{text\}:’”
\\

\bottomrule
\end{tabularx}
\caption{P-CoT5 task (Rhyme Word Generation) with sample prompts.}
\label{tab:5Peda-CoT-rhyme}
\end{table*}

\begin{table*}[t]
\centering
\footnotesize
\begin{tabularx}{\textwidth}{p{3cm} X}
\toprule
\textbf{Task} & \textbf{Prompt} \\
\midrule

\textbf{P-CoT5: G2P Conversion} 
& 
\textbf{System:}
You are a leading expert in General American English (GAE) phonology. 
Your specialty is converting written words (graphemes) into their precise GAE pronunciation (phonemes).

\textbf{User (example 1):}
“Let's analyze the word ‘apparently’ using standard GAE pronunciation. 
Show me its exact phonemic transcription, paying special attention to American rhotic sounds and stress patterns.”

\textbf{Assistant:}
“The General American English pronunciation of ‘apparently’ is 
\textipa{/@p"E\*r@ntli/}. 
In GAE, we have the unstressed initial \textipa{@p}, 
followed by the stressed syllable \textipa{"E\*r} with the characteristic American rhotic \textipa{\*r}, 
then the reduced \textipa{@nt}, and finally \textipa{li}.”

\textbf{User (example 2):}
“For our next GAE analysis, let's examine the word ‘calorie.’ 
What is its precise General American English phonemic transcription? 
Remember to account for the distinctive r-colored vowel that characterizes GAE.”

\textbf{Assistant:}
“The General American English pronunciation of ‘calorie’ is 
\textipa{/"k{\ae}l\textrhookschwa i/}. 
In GAE, we have the stressed syllable \textipa{"k{\ae}l}, 
followed by the r-colored schwa \textipa{\textrhookschwa}, 
and ending with \textipa{i}.”

\textbf{User (example 3):}
“Let's examine ‘freshman’ in General American English. 
What is its precise GAE phonemic transcription? 
Note particularly how the rhotic sound manifests in American pronunciation.”

\textbf{Assistant:}
“The General American English pronunciation of ‘freshman’ is 
\textipa{/"f\*rESm@n/}. 
In GAE, we begin with the consonant cluster \textipa{f\*r} featuring the American rhotic, 
followed by the stressed \textipa{"E S}, and ending with the reduced syllable \textipa{m@n}.”

\textbf{User (example 4):}
“Let's examine the word ‘breeze’ in General American English. 
What is its precise GAE phonemic transcription? 
Pay attention to the rhotic sound and vowel length.”

\textbf{Assistant:}
“The General American English pronunciation of ‘breeze’ is 
\textipa{/"b\*ri:z/}. 
In GAE, we have the initial consonant cluster \textipa{b\*r}, 
followed by the long vowel \textipa{i:z}, and ending with \textipa{z}.”

\textbf{User (example 5):}
“Now let's analyze the word ‘invite’ in General American English. 
What is its precise GAE phonemic transcription? 
Focus on the vowel quality and stress pattern.”

\textbf{Assistant:}
“The General American English pronunciation of ‘invite’ is 
\textipa{/In"vaIt/}. 
In GAE, we begin with the unstressed syllable \textipa{In}, 
followed by the stressed syllable \textipa{"vaIt} containing the diphthong \textipa{aI}.”

\textbf{User (final request):}
“Now, provide the precise General American English phonemic transcription for ‘\{text\}.’ 
Focus on accurate representation of American vowels, rhotics (\textipa{\*r}), and stress patterns. 
Give me \emph{only} its complete GAE phonemic transcription.”
\\

\bottomrule
\end{tabularx}
\caption{P-CoT5 task (G2P Conversion) with sample prompts.}
\label{tab:5Peda-CoT-G2P}
\end{table*}

\begin{table*}[t]
\centering
\footnotesize
\begin{tabularx}{\textwidth}{p{3cm} X}
\toprule
\textbf{Task} & \textbf{Prompt} \\
\midrule

\textbf{P-CoT5: Syllable Counting}
&
\textbf{System:}
You are an expert in American English phonology and phonetics. 
A syllable is a unit of pronunciation having exactly one vowel sound, forming the whole or a part of a word. 
Your goal is to assist the student in counting the total number of syllables in a given sentence using discovery learning methods.

\textbf{User:}
“I'm ready to learn about syllables and how to count them in a sentence.”

\textbf{Assistant:}
“Great! Let's start by understanding what a syllable is. 
Remember, a syllable has one vowel sound. 
Now let's try an example together.”

\textbf{User (example 1):}
“Sentence: ‘Grace has resigned herself to simply completing the upbringing of her teenage daughter.’ 
How many syllables do you think it has?”

\textbf{Assistant:}
“Let's break it down together. Start by identifying the vowel sounds in each word.”

\textbf{User:}
“Grace (1), has (1), resigned (2), herself (2), to (1), simply (2), completing (3), the (1), upbringing (3), of (1), her (1), teenage (2), daughter (2)”

\textbf{Assistant:}
“Exactly! Now, let's sum them all up.”

\textbf{User:}
“22”

\textbf{Assistant:}
“That's correct! There are 22 syllables in that sentence.”

\textbf{User (example 2):}
“Sentence: ‘This story is about a young girl's redemption in a small town.’ 
How many syllables do you think it has?”

\textbf{Assistant:}
“Let's work through this one as well. Identify the vowel sounds in each word.”

\textbf{User:}
“This (1), story (2), is (1), about (2), a (1), young (1), girl's (1), redemption (3), in (1), a (1), small (1), town (1)”

\textbf{Assistant:}
“Good job! Now add them together.”

\textbf{User:}
“16”

\textbf{Assistant:}
“Correct again! It has 16 syllables.”

\textbf{User (example 3):}
“Sentence: ‘The one thing that hasn't happened is a proposal.’ 
How many syllables do you think it has?”

\textbf{Assistant:}
“Let's do the same process and find the vowel sounds in each word.”

\textbf{User:}
“The (1), one (1), thing (1), that (1), hasn’t (2), happened (2), is (1), a (1), proposal (3)”

\textbf{Assistant:}
“Perfect! Now, sum them up.”

\textbf{User:}
“13”

\textbf{Assistant:}
“Well done! The total is 13 syllables.”

\textbf{User (example 4):}
“Sentence: ‘She meets him randomly in the woods at his family's cabin.’ 
How many syllables do you think it has?”

\textbf{Assistant:}
“Let's break it down by identifying the vowel sounds in each word.”

\textbf{User:}
“She (1), meets (1), him (1), randomly (3), in (1), the (1), woods (1), at (1), his (1), family's (3), cabin (2)”

\textbf{Assistant:}
“Good! Now, tally them up.”

\textbf{User:}
“16”

\textbf{Assistant:}
“Yes, that's 16 syllables.”

\textbf{User (example 5):}
“Sentence: ‘Just a simple blacksmith’s assistant, he didn’t have much to offer, but his love.’ 
How many syllables do you think it has?”

\textbf{Assistant:}
“Follow the same process and find the vowel sounds in each word.”

\textbf{User:}
“Just (1), a (1), simple (2), blacksmith's (2), assistant (3), he (1), didn’t (2), have (1), much (1), to (1), offer (2), but (1), his (1), love (1)”

\textbf{Assistant:}
“Nice work! Now add them up.”

\textbf{User:}
“20”

\textbf{Assistant:}
“Correct! The total number is 20 syllables.”

\textbf{User (final request):}
“Now it's your turn! Count the \emph{total} number of syllables in the given word (in Target Sentence): 
‘\{text\}.’ 
Do not randomly generate the number; rather, you must analyze the vowel sounds and sum them accurately. 
Provide \emph{only} the total count.”
\\

\bottomrule
\end{tabularx}
\caption{P-CoT5 task (Syllable Counting) with sample prompts.}
\label{tab:5Peda-CoT-syllable}
\end{table*}

\end{document}